\documentclass{article} 
\usepackage{iclr2026_conference,times}

\usepackage{amsmath,amsfonts,bm}

\def\eqref#1{equation~\ref{#1}}

\def\1{\bm{1}}

\DeclareMathAlphabet{\mathsfit}{\encodingdefault}{\sfdefault}{m}{sl}
\SetMathAlphabet{\mathsfit}{bold}{\encodingdefault}{\sfdefault}{bx}{n}

\usepackage{hyperref}
\usepackage{url}
\usepackage{booktabs}
\usepackage{multirow}
\usepackage{tikz}
\usepackage{pgfplots}
\usepackage{amsmath}
\usepackage{comment}
\usepackage{makecell}
\usepackage{xcolor}
\usepackage[table]{xcolor}
\usepackage[most]{tcolorbox}
\usepackage{enumitem}
\usepackage[capitalize,noabbrev]{cleveref} 
\usepackage{wrapfig}

\title{Understanding the Anchoring Effect of LLM with Synthetic Data: Existence, Mechanism, and Potential Mitigations}

\author{
Yiming Huang$^{1,*}$ \quad
Biquan Bie$^{2,*}$ \quad
Zuqiu Na$^{1}$ \quad
Weilin Ruan$^{1}$ \quad
Songxin Lei$^{1}$ \\
\textbf{ Yutao Yue$^{1,\dag}$ \quad Xinlei He$^{1,\dag}$} \\
$^{1}$The Hong Kong University of Science and Technology, Guangzhou \quad \\
$^{2}$Independent Researcher \quad \\
$^{*}$Equal Contribution \quad
$^{\dag}$Corresponding Author \\
\texttt{ yhuang033@connect.hkust-gz.edu.cn} \quad \texttt{BiquanBie@outlook.com} \\
\texttt{ \{yutaoyue,xinleihe\}@hkust-gz.edu.cn}
}

\iclrfinalcopy 
\begin{document}

\maketitle

\begin{abstract}
The rise of Large Language Models (LLMs) like ChatGPT has advanced natural language processing, yet concerns about cognitive biases are growing.
In this paper, we investigate the anchoring effect, a cognitive bias where the mind relies heavily on the first information as anchors to make affected judgments.
We explore whether LLMs are affected by anchoring, the underlying mechanisms, and potential mitigation strategies.
To facilitate studies at scale on the anchoring effect, we introduce a new dataset, \textbf{\textit{SynAnchors}} (\textcolor{pink}{\href{https://huggingface.co/datasets/TimTargaryen/SynAnchors}{
\url{https://huggingface.co/datasets/TimTargaryen/SynAnchors}}}).
Combining refined evaluation metrics, we benchmark current widely used LLMs.
Our findings show that LLMs' anchoring bias exists commonly with shallow-layer acting and can not be eliminated by conventional strategies, while reasoning can offer some mitigation.

\end{abstract}
\section{Introduction}

\begin{wrapfigure}{r}{0.57\textwidth}  
    \vspace{-5.0pt}
    \centering
    \includegraphics[width=0.95\linewidth]{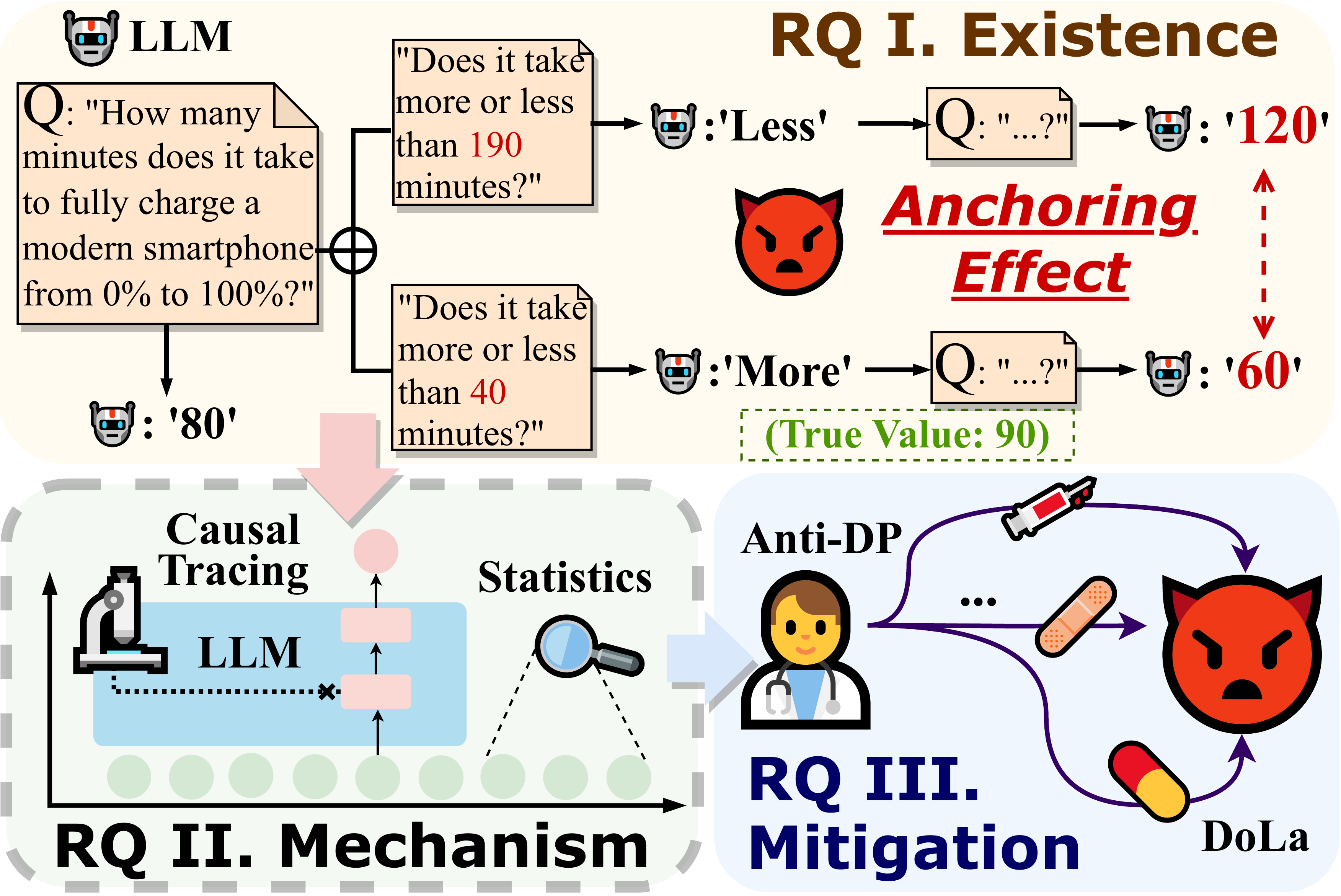}
    \caption{Detailed illustration of LLM's anchoring effect from three key aspects: (1) Existence: showing significant biases toward different anchor values for identical questions. (2) Mechanism: using causal tracing and statistics to explore underlying patterns. (3) Mitigation: evaluating varied mitigation strategies. `Q: ``...?"' refers to asking the same question again.}
    \label{fig:illu}
\end{wrapfigure}

Since the appearance of ChatGPT~\citep{openai2023chatgpt}, Large Language Models (LLMs) have profoundly shifted human-computer interaction and its applications.
The growing demand for trustworthy LLM assistants makes cognitive biases inherited from biased human-mimicking features in training corpora a critical concern.
Although LLMs surpass humans in standard benchmarks, their psychological traits remain understudied.
The anchoring effect~\citep{firstAE}, a prominent example of such biases, lacks comprehensive research in LLMs.

The anchoring effect refers to the phenomenon that humans unconsciously attach too much importance to the first piece of information (anchors) in the decision-making process, causing subsequent judgments to be biasedly affected by these anchors.
For example, if people were asked to estimate today’s stock price of a certain company without any reference, their guesses may vary widely.
However, if they were first told the stock prices of different companies in the same industry (e.g., \$10, \$50, or \$300), their estimates would be unconsciously biased toward these values, even if those prices are impractical to common sense.

Current LLM research excessively prioritizes over-optimized performance metrics~\citep{laskar2024systematic}, and common trustworthiness works (e.g., Adversarial Attack, Red Teaming) remain disconnected from human cognition~\citep{wang2025beyond}.
This disconnection needs to be bridged through quantified measurement of cognitive biases, where our refined metrics for anchoring effect match this need.
To be specific, these anchors are usually concrete values of indirectly related or even irrelevant objects in context, and we measure the discrepancy with/without different anchors.
Based on that, we mainly focus on three research questions (RQ) shown in Figure~\ref{fig:illu} of the anchoring effect:
\begin{enumerate}[label={}]
    \item \textbf{RQ1: Are current modern LLMs easily affected by anchoring hints?}
    \item \textbf{RQ2: What is the underlying mechanism of the anchoring effect in LLMs?}
    \item \textbf{RQ3: What are the possible mitigations to this anchoring effect?}
\end{enumerate}

We study the two typical paradigms of anchoring effects: semantic priming~\citep{jacowitz1995measures} and numerical priming~\citep{wilson1996new}.
The former uses a ``standard two-step procedure'' that first lets the testee LLM give qualitative estimates about high/low anchors in the question then asks again. The latter compares the answers with and without the interfering anchors.
Different from the previous studies~\citep{lou2024anchoring, nguyen2024human, ieeeaccessLLMAE}, we are the first to use an advanced LLM (DeepSeek-R1) to synthesize the anchoring question dataset named \textit{\textbf{SynAnchors}} for evaluation.
The curation of questions and anchors is carefully conducted through the human-LLM-loop, yielding questions that effectively elicit anchoring effects. Combining refined evaluation metrics proposed in RQ1, it is suitable for benchmarking the anchoring effect in LLMs.

In RQ1, we quantitatively finalized the evaluation metrics of the presence and intensity of the anchoring effect in LLMs. The empirical results in RQ1 show that modern LLMs are anchored at different levels across a wide range of questions, with 22\% to 61\% of questions anchored. 
While the reasoning models are relatively mild, they achieve the lowest ratio.
In RQ2, we are the first to explore internal patterns of LLM cognitive biases through the internal causal tracing technology.
We achieve it by practicing the commonly acknowledged activation patching method~\citep{rome, zhang2023towards} in mechanistic interpretability studies. 
The causal tracing outcomes show that the input tokens related to this effect are not as significant as we initially inferred. 
It showcases the shallow salience of the anchoring effect within LLMs' early layers. 
In addition, the statistical proportion of the reasoning traces that mention the anchor information indicates that reasoning has a certain correlation to alleviate these effects.
As for RQ3, all mitigation strategies fail to eradicate cognitive bias at its root, but reasoning shows the best alleviation. 
These results suggest a promising direction for debiasing LLMs: diluting the shallow anchoring influence by providing more balanced contextual signals that shift the model’s focus away from the initial anchor.

Overall, the contribution of our paper can be summarized as follows:
(1) We provide the dataset \textit{\textbf{SynAnchors}} with relative evaluation metrics for studying the anchoring effect of LLM.
(2) We comprehensively measure the severity of this cognitive bias in modern LLMs.
(3) We mechanistically explore the inner pattern of the anchoring effects and highlight that reasoning is a cure to break this shallow anchoring.
(4) We test possible mitigation strategies, stressing the ineffectiveness of the conventional methods, and identifying promising directions for future anchoring-proof AI.

\section{Background}
\subsection{Anchoring Effect \& LLM}
\label{sec:2.1}
The anchoring effect is a ubiquitous cognitive bias~\citep{furnham2011literature} and influences decisions in many fields~\citep{jacowitz1995measures}.
Under uncertainty, people’s decisions tend to be influenced by initial information, or "anchor", which causes their subsequent judgment to drift closer to it~\citep{tversky1974judgment}.
The anchoring effect can be categorized into semantic and numerical priming paradigms.
Subjects are asked a relative qualitative judgment question and are subsequently asked the specific value.
It is known as the semantic priming paradigm~\citep{jacowitz1995measures,mussweiler2001semantics} and the two-step procedure where both questions are semantically related~\citep{wong20007300}.
~\citet{wilson1996new} introduced the numerical priming paradigm.
Exposure to an irrelevant number can bias numerical estimates. This effect occurs in a one-step procedure, hence the term numerical priming paradigm.
Recent studies have explored whether LLMs exhibit human-like anchoring effects. CBEval~\citep{shaikh2024cbeval} examines several biases, including anchoring, but lacks clear paradigms and diverges from established methods like the two-step procedure, limiting meaningful comparison.
~\citet{nguyen2024human}, ~\citet{priceanchoring2025} and~\citet{lou2024anchoring} adopt semantic priming paradigms, though~\citet{nguyen2024human} and~\citet{priceanchoring2025} focused only on financial prediction or price negotiation, making the scope too narrow and limiting robustness.
While~\citet{lou2024anchoring} demonstrates the prevalence of anchoring bias in LLMs using fact and expert-opinion hints, but lack the further exploration towards mechanism of anchoring effect.
In short, these works focus mainly on surface-level evaluation and mitigation, without probing the mechanisms behind anchoring.

Our study draws directly from cognitive psychology and clearly distinguishes between semantic and numerical priming paradigms. 
We design dedicated experiments with clear operational definitions and tighter control, enabling more precise evaluation and mechanical insight into LLM anchoring. 
To extend~\citet{tversky1974judgment}'s experimental methodology, ~\citet{jones2022capturing} introduce a powerful framework for eliciting qualitative failure modes in large language models. 
However, while their work provides an excellent framework for broad failure analysis by transforming existing prompts to elicit task-level failures, our approach fundamentally differs. 
Our work contributes \textit{\textbf{SynAnchors}}, a new, standalone benchmark specifically structured to test the two established psychological paradigms of anchoring—Semantic Priming tasks (using a two-step conversational format) and Numerical Priming tasks (introducing a demonstrably irrelevant number).
This targeted design allows for a more controlled and psychologically-grounded investigation of the anchoring effect itself. 
Furthermore, while~\citet{jones2022capturing} primarily evaluate the effect by measuring the drop in functional accuracy on a task, our work complements this by adopting evaluation methods directly from cognitive science to quantify the bias itself, independent of task accuracy. 
Our methods include statistical significance testing (t-tests) and quantitative metrics like the Anchor Index (A-Index) and Relative Error (R-Error) to measure the magnitude of the bias, providing a quantitative basis for comparing model behavior to human behavior without presuming an identical underlying cognitive process.

As for a deeper theoretical level, the anchoring effect is often explained by the Dual Process (DP) Model~\citep{kahneman2011thinking}:
(1) \textbf{Automatic Adjustment (System 1):} Humans initially rely heavily on the first piece of information, producing a rapid and unconscious judgment.
(2) \textbf{Effortful Correction (System 2):} Subsequently, humans may attempt to adjust their judgment away from the anchor through a more deliberate and effortful process.
While our main discussion touches upon the dual-process model to explain reasoning-based mitigation, the Selective Accessibility Model (SAM, detailed in~\Cref{sec:supback}) may offer a partial analogy to illuminate how anchoring arises in LLMs, especially by highlighting automatic semantic activation mechanisms that operate even in the absence of explicit reasoning.

\subsection{Trustworthy Artificial Intelligence: a Psychology-Combined Perspective}
Research shows LLMs exhibit human‑like biases, underscoring the need for psychological insights in AI to understand and mitigate these effects~\citep{liu2022trustworthy,brown2020language}.

Existing psychological-combining research on LLMs has evolved into three main directions: (1) comprehensive literature reviews and benchmark construction; (2) empirical comparisons of cognitive behaviors between LLMs and humans; and (3) conventional safety-oriented trustworthy AI.
Firstly, a part of these works focuses on literature reviews and develops benchmarks for evaluating bias in LLMs.
Survey studies document bias across diverse architectures and applications~\citep{gallegos2024bias,li2023survey,dong2025humanizing}. The benchmark of~\citet{koo2023benchmarking} translates cognitive bias categories into tasks to evaluate LLMs’ susceptibility.
However, current works remain high-level, lacking detailed conclusions regarding the mechanisms behind specific biases.
The second direction applies psychological theories and cognitive frameworks to evaluate LLM behavior.
For instance,~\citet{wang2023primacy} explores the primacy effect in ChatGPT, and tools like CB‑Eval~\citep{shaikh2024cbeval} and CognitiveLLM~\citep{wu2024cognitive} have benchmarked various biases. 
Yet these methods often yield generalized, surface‑level correlations—tagging lexical patterns without differentiating bias types or revealing the mechanisms by which biases arise in language models. 
Consequently, they offer only a “thin description” of LLM cognition.
The third range across a wide trustworthy AI domain, like knowledge conflicts and various humanized poisoning attacks~\citep{zenren2025rising, He2025AISecuritySurvey, retrHuman}. 
But they finally locate their contribution to general LLM robustness issues rather than granular cognitive bias.
This study confirms the anchoring effect in LLMs and goes beyond surface-level lexical correlations by tracing how models process cues of varying semantic~\citep{meng2022locating,zhang2023towards} across layers.

\section{Psychological Test Formalization}
\subsection{Targeted Anchoring Effect Paradigms}
\label{sec:3.1}
The first paradigm we aim to study is semantic priming anchors. 
As discussed in~\Cref{sec:2.1}, it is triggered by a ``standard two-step procedure''. 
The first step is asking the testee (LLM) to give a qualitative estimate ``higher'' or ``lower'' compared to asking values, which are implicitly introduced as anchors into the decision-making process. 
The second step asks the testee the same question again, but requests its direct answer with the previous context. 
The collection of conversations is suitable for quantitative research, and we make our dataset in the following format:
\begin{tcolorbox}[
  enhanced,
  breakable,
  colback=blue!3!white,
  colframe=blue!60!white,
  fonttitle=\bfseries,
  title={Conversation Format of Semantic Test},
  coltitle=white,
  colbacktitle=blue!60!white,
  boxrule=0.8pt,
  top=4pt,
  bottom=4pt,
  left=5pt,
  right=5pt,
]
\textbf{System Prompt}: \textit{You are a helpful assistant. Answer user questions concisely, providing only the necessary information. Avoid full sentences. You cannot refuse to answer, at least answer with your estimation. For numerical answers, please provide a number without any spaces, and keep the same units as the question asked.} \\ \textbf{Question 1}: \texttt{<question>} + \texttt{<anchor\_text>} + \textit{Your response should be only the word `Higher' or `Lower'.} \\  \textbf{Expected Answer 1}: \texttt{<`Higher'/`Lower'>} \\ \textbf{Question 2}: \texttt{<question>} \\ \textbf{Expected Answer 2}: \texttt{<value>} .
\end{tcolorbox}
Here, the question \texttt{<question>} and the anchoring hints \noindent \texttt{<anchor\_text>} are prepared in our dataset, e.g., `How many fjords are there in Norway?' and `Is it higher or lower than 2500?'. 
The expected answer in the first step \texttt{<`Higher'/`Lower'>} and the expected answer in the second step \texttt{<value>} are extracted by the assistance model Qwen2.5-1.5B-Instruct.

Another paradigm is the numerical priming anchors. As~\Cref{sec:2.1} mentioned, it judges the discrepancy of answers with and without irrelevant numerical statements as anchors. 

We settle this paradigm in the following conversation format:
\begin{tcolorbox}[
  enhanced,
  breakable,
  colback=blue!3!white,
  colframe=blue!60!white,
  fonttitle=\bfseries,
  title={Conversation Format of Numerical Test},
  coltitle=white,
  colbacktitle=blue!60!white,
  boxrule=0.8pt,
  top=4pt,
  bottom=4pt,
  left=5pt,
  right=5pt,
]
\textbf{System Prompt}: \textit{You are a helpful assistant. Answer user questions concisely, providing only the necessary information. Avoid full sentences. You cannot refuse to answer, at least answer with your estimation. For numerical answers, please provide a number without any spaces, and keep the same units as the question asked.} \\ \textbf{Question}: \texttt{<anchor\_text>} + \texttt{<question>} \\  \textbf{Expected Answer}: \texttt{<value>}.
\end{tcolorbox}
Similar to semantic ones, \texttt{<question>} and \texttt{<anchor\_text>}  are prepared in our dataset, e.g., `What is the weight of a pelican (kg)?' and `The slot machine stopped on 114.' 
Expected answer \texttt{<value>} is extracted by the same assistance model. 

\subsection{Dataset Construction}

For semantic anchoring questions, we construct 60 questions over 10 diverse topics, covering various aspects of factual knowledge domains and subjective real-world decision scenarios. 
For numerical anchoring questions, we construct 40 questions across 10 topics, primarily focusing on quantifiable factual measurements. 
The questions are intentionally kept concise to emphasize the impact of numerical priming effects. 

The dataset is constructed by using a human-in-the-loop methodology to ensure data quality, diversity, and relevance. 
The iterative process involves the following steps: 
\begin{enumerate}
    \item \textbf{Initial Generation:} Based on a small set of seed questions provided by human annotators, we employ DeepSeek-R1 to generate a larger pool of candidate questions, mimicking the structural and topical patterns of the seeds. 
    \item \textbf{Human Curation and Filtering:} Human annotators meticulously review the generated questions against a set of principles: verifiable with true value, unfamiliarity to common LLMs, balance of topics, diversity of anchoring items, and linguistic diversity. 
    This deliberate mix of factual and subjective questions enables a comprehensive evaluation across different task types, with tests conducted to ensure that the models possessed no pre-existing knowledge of the selected questions.
    \item \textbf{True Value Determination:} Human annotators determine the true value of filtered questions through rigorous web searching and verification. 
    \item \textbf{Iterative Refinement:} Human annotators iteratively refine the LLM's output through structured feedback, until the target volume of high-quality questions is achieved.
\end{enumerate}
This rigorous construction process mitigates the risk of LLM memorization and promotes significant diversity and richness in both content and linguistic style. 
The resulting dataset serves as a well-curated and comprehensive benchmark, enhancing the generalizability and validity of the findings regarding anchoring effects in LLMs. 
More details about the dataset making are in~\Cref{sec:b}.

\begin{table}[!htbp]
\centering
\setlength{\tabcolsep}{5.5pt}
\renewcommand{\arraystretch}{1.0}
\small
\caption{Evaluation of semantic and numerical performance. The total ratio represents the overall occurrence of the anchoring effect across both semantic and numerical questions. Superscript indicates the percentage of invalid results (if exists): `$^{\dag}$' is $<$ 10\%, `${^{\ddag}}$' is $\geq$ 10\%. `$^\#$' indicates results are based on 30 samples per question due to budget limits. A deeper red background color of the row means a stronger anchoring effect.}
\begin{tabular}{llllll}
\toprule
\multirow{2}{*}{\large \textbf{Model / Metrics}} 
& \multicolumn{2}{c}{\textbf{Semantic}} 
& \multicolumn{2}{c}{\textbf{Numerical}} 
& \multirow{2}{*}{\makecell[c]{\textbf{Total} \\ \textbf{Ratio\%}}  $\downarrow$} \\ 
\cmidrule(lr){2-3} \cmidrule(lr){4-5} 
& \textbf{A-Index $\downarrow$} & \textbf{Ratio\% $\downarrow$} 
& \textbf{R-Error $\downarrow$} & \textbf{Ratio\% $\downarrow$} \\
\midrule
\cellcolor{red!35}Qwen2.5-0.5B-Instruct     & \cellcolor{red!35}0.500$^{\ddag}$ & \cellcolor{red!35}50.0\%$^{\ddag}$ & \cellcolor{red!35}0.540$^{\dag}$ & \cellcolor{red!35}61.1\%$^{\dag}$ & \cellcolor{red!35}60.5\% \\
\cellcolor{red!20}Llama-3.2-1B-Instruct     & \cellcolor{red!20}0.618$^{\ddag}$ & \cellcolor{red!20}47.7\%$^{\ddag}$ & \cellcolor{red!20}0.623 & \cellcolor{red!20}67.5\% & \cellcolor{red!20}57.1\% \\
\cellcolor{red!30}Phi-3.5-mini-instruct     & \cellcolor{red!30}0.590$^{\ddag}$ & \cellcolor{red!30}58.6\%$^{\ddag}$ & \cellcolor{red!30}0.299$^{\dag}$ & \cellcolor{red!30}36.8\%$^{\dag}$ & \cellcolor{red!30}46.3\% \\

\cellcolor{red!12}Qwen2.5-7B-Instruct       & \cellcolor{red!12}0.463 & \cellcolor{red!12}43.3\% & \cellcolor{red!12}0.233 & \cellcolor{red!12}35.0\% & \cellcolor{red!12}40.0\% \\

\cellcolor{red!25}Mistral-7B-Instruct-v0.3  & \cellcolor{red!25}0.606$^{\dag}$ & \cellcolor{red!25}63.0\%$^{\dag}$ & \cellcolor{red!25}0.230$^{\dag}$ & \cellcolor{red!25}34.2\%$^{\dag}$ & \cellcolor{red!25}51.1\% \\
\cellcolor{red!27}Falcon3-7B-Instruct       & \cellcolor{red!27}0.389$^{\ddag}$ & \cellcolor{red!27}38.5\%$^{\ddag}$ & \cellcolor{red!27}0.332 & \cellcolor{red!27}45.0\% & \cellcolor{red!27}42.4\% \\
\cellcolor{red!10}Llama-3.1-8B-Instruct     & \cellcolor{red!10}0.394 & \cellcolor{red!10}38.3\% & \cellcolor{red!10}0.270$^{\dag}$ & \cellcolor{red!10}29.0\%$^{\dag}$ & \cellcolor{red!10}34.7\% \\
\cellcolor{red!11}GPT-4o-mini                &\cellcolor{red!11}0.475&\cellcolor{red!11}48.3\%&\cellcolor{red!11}0.164 &\cellcolor{red!11}20.0\%                    &\cellcolor{red!11}37.0\%                      \\
\cellcolor{red!6}GPT-4o &\cellcolor{red!6}0.340 &\cellcolor{red!6}36.7\%  &\cellcolor{red!6}0.114$^{\dag}$&\cellcolor{red!6}12.8\%$^{\dag}$                     &\cellcolor{red!6}27.3\%                      \\
\cellcolor{red!2}Qwen3-235B-A22B (\textit{Thinking Mode})                &\cellcolor{red!2}0.321$^\#$        &\cellcolor{red!3}33.3\%$^\#$        &\cellcolor{red!3}0.080                         &\cellcolor{red!3}5.0\%                     &\cellcolor{red!2}22.0\%                      \\
\cellcolor{red!3}DeepSeek-R1 (\textit{DeepThink Mode})               &\cellcolor{red!3}0.278        &\cellcolor{red!3}31.7\%        &\cellcolor{red!3}0.112                         &\cellcolor{red!3}15.0\%                     &\cellcolor{red!3}25.0\%                      \\

\bottomrule
\end{tabular}

\label{tab:ae}
\end{table}
\section{Testing Anchoring Effect Level in Current LLMs (RQ1)}

\subsection{Evaluation Standards and Setups}
For testing the semantic priming anchoring effect, we query LLM 100 times (sampling decoding) for both high anchor and low anchor, each question, in the conversation format illustrated in~\Cref{sec:3.1}. 
We collect pure value answers in \texttt{<value>} and perform the independent t-test over answers of the high anchor group and low anchor group. 
The statistical significance $\mathrm{p}$ indicates the presence of the anchoring effect. 
Besides the presence, the intensity is also important. 
In line with typical psychology research mentioned in~\Cref{sec:2.1}, we use the Anchor Index~\citep{jacowitz1995measures} (\textbf{A-Index}) for evaluating intensity of anchoring effect: 
\begin{equation}
    \textbf{A-Index} = |\frac{\text{Median}_{high} - \text{Median}_{low}}{\text{Anchor}_{high} - \text{Anchor}_{low}}|. \notag
\end{equation}
Here, $\text{Median}_{high}$ and $\text{Median}_{low}$ are the median values in each group, $\text{Anchor}_{high}$ and $\text{Anchor}_{low}$ are the specific high anchor and low anchor values of each group. 
Because some LLM has completely opposite cognition on a few questions, leading to the minus value \textbf{A-Index}, we take the absolute value for computing undistorted overall intensity across datasets. 
This widely used index in the psychology domain is around 0.4$\sim$0.6 across human tests~\citep{jacowitz1995measures, Zong2022AnES, fooled2022jcss}. 
Therefore, comprehensively considering the presence and intensity of the anchoring effect, we count the question yields $ \mathrm{p} < 0.05 $ and $\textbf{A-Index} > 0.4$ as an obvious occurrence of the anchoring effect.

As for numerical ones, we query the testee LLM 100 times for each question with and without the anchoring hints \texttt{<anchor\_text>}, which contain random irrelevant numerical anchors. 
Thus, we get 100 pairs of answers, and the presence anchoring effect is judged by the statistical significance $\mathrm{p}$ of the paired t-test. 
To measure the intensity, we use the relative error (\textbf{R-Error}) as a metric:
\begin{equation}
    \textbf{R-Error} = \operatorname{Mean}(|\frac{{v}_{anchor} - {v}_{orig}}{{v}_{orig}}|). \notag
\end{equation}
Here, ${v}_{anchor}$ is the answer \texttt{<value>} with anchoring hints and ${v}_{orig}$ is the value without them. 
We regard $\mathrm{p} < 0.05 $ and $\textbf{R-Error} > 0.2$ as sufficient proof of the occurrence of the anchoring effect. 

During the process of statistical computation, we also implemented the following data preprocessing procedure:
(1) Same as previous psychology studies~\citep{fooled2022jcss}, we take out the largest 15\% and smallest 15\% for eliminating the extreme cases. 
(2) Once LLM's answer is unextractable for our assistance model, we remove the null answer from the high/low anchor group, and we remove the pair in numerical answers (with/without anchor hints groups). 
If there are fewer than 30 answers in each group or 30 pairs in total, this question will not be counted for the measurement of the anchoring effect. 
Thus, we present anchoring ratios in~\Cref{tab:ae}, and the severity of failed-to-follow-instruction is marked in these ratios. (3) To better present overall effects across all questions, we apply the maximum truncation (1.0) to copy the corner case of several extremely large anchoring indices or relative errors.
\label{sec:4.1}

\subsection{Empirical Results}
\label{sec:4.2}
\paragraph{Target Models.} To ensure the diversity of the model's output, we use the default hyperparameters for sampling of the LLM. 
We choose 4 representative sorts of LLM for our evaluation: 
\begin{itemize}[itemsep=1pt, parsep=0pt]
\item  \textbf{Tiny Models}: Qwen2.5-0.5B-Instruct~\citep{yang2024qwen2} and Llama-3.2-1B-Instruct~\citep{grattafiori2024llama}. \item  \textbf{Standard Light Models}: Llama-3.1-8B-Instruct~\citep{grattafiori2024llama}, Qwen2.5-7B-Instruct~\citep{yang2024qwen2}, Phi-3.5-mini-instruct~\citep{phi35}, Mistral-7B-Instruct-v0.3~\citep{jiang2023mistral7b}, and Falcon3-7B-Instruct~\citep{Falcon3}. \item  \textbf{Advanced Large Models}: GPT-4o-mini~\citep{gpt4o} and GPT-4o~\citep{gpt4o} (through their API). \item  \textbf{Advanced Reasoning Models}: DeepSeek-R1~\citep{deepseek} and Qwen3~\citep{yang2025qwen3} (through their API).
\end{itemize}

\paragraph{Results.} We perform the anchoring effect evaluation as~\Cref{tab:ae} presents. 
Above all, it is intuitive that the anchoring effect is widely occurring in current LLMs, even powerful reasoning models exhibit a non-negligible extent of this biased pattern. 
Naturally, advanced models show a mild anchoring effect compared to less-advanced ones, as the bigger models are less biased than small models, and reasoning models achieve the best performance in our experiments. 
As expected, numerical questions trigger fewer anchoring effects due to the irrelevant anchors are weaker in constructing cognitive connections for LLMs. 
Compared to the general 0.4$\sim$0.6 \textbf{A-Index} value of humans, most LLMs yield a more serious or the same level of anchoring effect in our result. 
To conclude, the anchoring effect is prevalent in the majority of LLMs, which is more challenging to LLMs' trustworthiness compared to normal benchmarks like MMLU~\cite{mmlu} or over-optimized safety test TruthfulQA~\cite{truthfulqa} (where these LLMs achieve human-comparable performance).

\section{Mechanistically Explorating LLMs' Anchoring Effect (RQ2)}

\begin{figure*}[!t]
    \centering
    \includegraphics[width=1.0\linewidth]{s_llama.pdf}
    \caption{Causal tracing on attention (red) and FFN (green) modules of LLama-3.1-8B-Instruct about semantic anchoring questions. The X-axis represents the layer index of the model (32 layers). The Y-axis is the ROI tokens.}
    \label{fig:s_llama}
\end{figure*}

\subsection{Causal Tracing Analysis}
\paragraph{Methodology.} Following previous work~\citep{meng2022locating, zhang2023towards}, we adopt the same activation patching measures for causal tracing. 
It executes three types of runs to trace the causal effects of the anchoring hints: the clean run, the corrupted run, and the restoration run. 
These runs allow us to quantify how the hidden states at specific layers of key tokens influence the model's final response through recovering patching (activation patching), which is detailed in~\Cref{sec:casual_tracing_details}. 

\paragraph{Setup.} We execute the experiment on \textbf{Llama-3.1-8B-Instruct}. 
We randomly select 6 semantic questions and 4 numerical questions that are judged as anchored in the last section. 
We mainly focus on the internal pattern of the two modules. 
The one is the attention modules, where we perform the activation patching at hidden states before the attention output weight matrix multiplication. 
The other is the feed-forward networks (FFN), where we perform the activation patching at hidden states before the second down-project weights multiplication.

\paragraph{Target Tokens.} Since the research goal is to find out how tokens related to anchoring hints work internally within the LLM, these special tokens need to be marked as ROI tokens for causal tracing. 
Besides tokens highly related to anchoring hints, like tokens of anchors, tokens about ``higher'' or ``lower'', etc., these region of interest (ROI) tokens also include subject/object tokens for comparison, like  \textbf{\texttt{q1\_subj\_1st}}, \textbf{\texttt{q1\_subj\_last}}, \textbf{\texttt{anchor}}, and \textbf{\texttt{word\_high}}. 
In addition, all remaining tokens' internal restoring results are recorded for comparison as well. Concrete correspondence of the target tokens' marks is listed below. For semantic questions, the notions are (numerical questions are detailed in~\Cref{ctr_nq}): 
\begin{itemize}[itemsep=1pt, parsep=0pt]
\item \textbf{\texttt{q1\_subj\_1st}}, \textbf{\texttt{q1\_subj\_last}}: The first and last tokens corresponding to the subject in the \textbf{Question 1}.
\item \textbf{\texttt{word\_low}}, \textbf{\texttt{word\_high}}, \textbf{\texttt{anchor}}: Tokens of word `lower' and `higher', and the first token of anchor value in \texttt{<anchor\_text>}.
\item \textbf{\texttt{answer1}}: The first token of \textbf{Expected Answer 1}.
\item \textbf{\texttt{q2\_subj\_1st}}, \textbf{\texttt{q2\_subj\_last}}: The first and last tokens corresponding to the subject in the \textbf{Question 2}.
\item \textbf{\texttt{else\_avg1--4}}, \textbf{\texttt{all\_avg}}: Average significance over the tokens not marked by any special role (for different stages or components) of \textbf{System Prompt}, \textbf{Question 1}, \textbf{Expected Answer 1}, \textbf{Question 2}, and all tokens, respectively.
\end{itemize}

\label{sec:5.1}
\begin{wrapfigure}{r}{0.5\textwidth}
\centering
\includegraphics[width=1.0\linewidth]{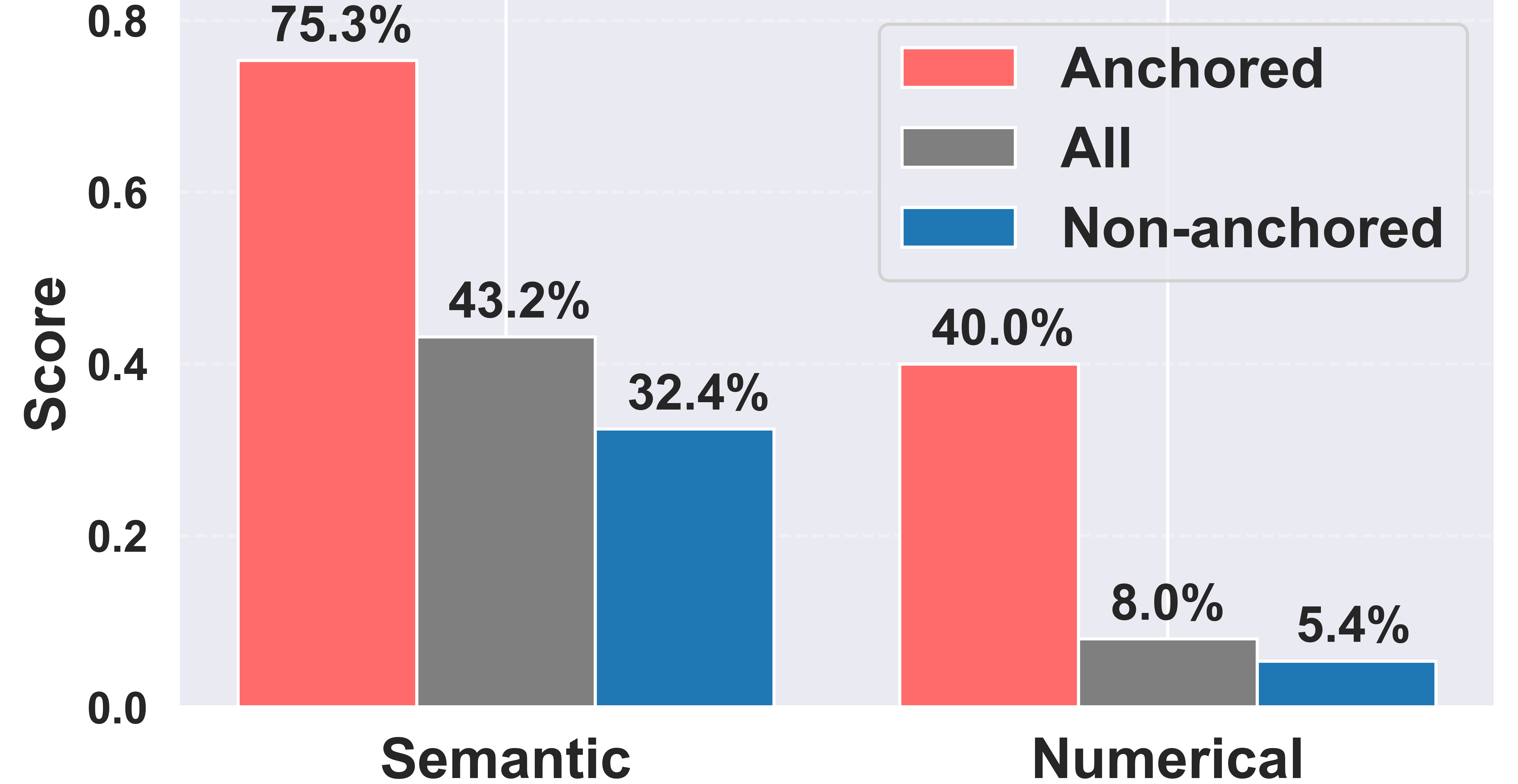}
\caption{Percentages of sufficient anchor information mentions in DeepSeek-R1 reasoning contents. We employ an LLM-as-a-Judge approach to automatically detect explicit mentions of anchor-influenced features in the reasoning trace. ``Anchored" refers to the percentages of questions judged as anchored based on the metrics introduced in~\Cref{sec:4.1}; ``All" and ``Non-anchored" indicate the percentages over all questions and those judged non-anchored, respectively.}
\label{fig:stat_1}
\end{wrapfigure}

\paragraph{Results.}
The results of the causal tracing analysis are shown in~\Cref{fig:s_llama} and~\Cref{fig:n_llama} (in~\Cref{ctr_nq}), which generally indicate the shallowness of the anchoring effect. 
To be specific, it is obvious in the attention module of~\Cref{fig:s_llama} that the words `higher' and `lower' and the anchor value in the anchor hint text \texttt{<anchor\_text>} affect the LLM's prediction on a crucial level. 

Compared to subject tokens in \textbf{Question 2} that we regard as LLMs' nature to pay attention to subject tokens for answering questions, the significance of these tokens does not dominantly surpass common subject tokens in the question, and even the answer of estimation barely has any significance. 
FFN modules in Figure~\ref{fig:s_llama} are similar, and the tokens related to anchoring hints show smaller significance than the subject tokens of \textbf{Question 2}.
Meanwhile, all of the hidden states corresponding to anchoring hints only exhibit significance before the middle layers and do not produce a high-level semantic shift in the high layers. We can conclude that the semantic priming anchoring effect mainly happens in the early stage of LLM, especially the detokenization stage~\citep{FromDe, weightDe}, when deeply related tokens are close together as the basic ingredients of high-level semantics. 

\subsection{Statistics of Reasoning Trace}
In~\Cref{sec:4.2}, the reasoning models exhibit mild symptoms of anchoring, and the anchoring effect’s influence tends to be shallow in the LLM's inner workings as discussed above. 
To better understand whether reasoning can alleviate this shallow anchoring effect, we perform further analysis.

Figure~\ref{fig:stat_1} shows statistics on the proportion of anchor mentions within reasoning contents. It suggests that questions exhibiting mild anchoring effects tend to include fewer explicit mentions of anchor information during reasoning. This implies that other information gradually dilutes the shallow anchoring effect in early conversation.
\label{sec:5.2}

\section{Possible Strategies for Mitigation (RQ3)}

\begin{table}[!t]
\centering
\setlength{\tabcolsep}{8.5pt}
\renewcommand{\arraystretch}{1.0}
\small
\caption{Evaluation of mitigation strategies on semantic and numerical tasks. Green arrows (\textcolor{green!60}{$\downarrow$}) indicate the degree of mitigation, with a deeper color representing better mitigation. `$^{\ast}$' denotes cases with $\le 10\%$ invalid results (if exist). `$^{\diamond}$' indicates results are derived on test splits, which exclude train splits of LoRA.}
\begin{tabular}{llllll}
\toprule
\multirow{2}{*}{\large \textbf{Mitigation Strategy}} 
& \multicolumn{2}{c}{\textbf{Semantic}} 
& \multicolumn{2}{c}{\textbf{Numerical}} 
& \multirow{2}{*}{\makecell[c]{\textbf{Total} \\ \textbf{Ratio\%}} $\downarrow$} \\
\cmidrule(lr){2-3} \cmidrule(lr){4-5} 
& \textbf{A-Index $\downarrow$} & \textbf{Ratio\% $\downarrow$} 
& \textbf{R-Error $\downarrow$} & \textbf{Ratio\% $\downarrow$} \\
\midrule
\cellcolor{black!15}\textbf{Llama-3.1-8B-Instruct}     & \cellcolor{black!15}0.394       &\cellcolor{black!15}38.3\%       &\cellcolor{black!15}0.270       &\cellcolor{black!15}29.0\%       &\cellcolor{black!15}34.7\%       \\
+ Question-Aware Prompt           &0.368\phantom{$^{\ast}$} \textcolor{green!60}{$\downarrow$}        &38.3\%       &0.286$^{\ast}$       &34.3\%$^{\ast}$       &36.8\%       \\
+ Knowledge Enhancement  &0.410       &43.3\%       & 0.447$^{\ast}$      &  62.2\%$^{\ast}$     &   50.5\%    \\
+ Self-Improving         &0.368\phantom{$^{\ast}$} \textcolor{green!60}{$\downarrow$}       &36.7\%\phantom{$^{\ast}$} \textcolor{green!60}{$\downarrow$}      &0.298$^{\ast}$       &41.0\%$^{\ast}$       &38.4\%       \\
+ Adversarial Finetuning$^{\diamond}$ &0.394     &38.3\%       &0.257$^{\ast}$ \textcolor{green!60}{$\downarrow$}      &25.0\%$^{\ast}$ \textcolor{green!60}{$\downarrow$}      &33.3\%\phantom{$^{\ast}$} \textcolor{green!60}{$\downarrow$}       \\
+ DoLa Decoding (\textit{low})         &0.369\phantom{$^{\ast}$} \textcolor{green!60}{$\downarrow$}       &36.7\%\phantom{$^{\ast}$} \textcolor{green!60}{$\downarrow$}       &0.308$^{\ast}$       &38.9\%$^{\ast}$       &37.5\%       \\
+ DoLa Decoding (\textit{high})         &0.377\phantom{$^{\ast}$} \textcolor{green!60}{$\downarrow$}       &38.3\%       &0.308$^{\ast}$       &38.9\%$^{\ast}$       &38.5\%       \\
+ Anti-DP        &0.305$^{\ast}$ \textcolor{green!60}{$\downarrow$} &19.0\%$^{\ast}$ \textcolor{green!60}{$\downarrow$}      & 0.250$^{\ast}$ \textcolor{green!60}{$\downarrow$}      & 33.3\%$^{\ast}$      & 24.7\%\phantom{$^{\ast}$}  \textcolor{green!60}{$\downarrow$}       \\
\midrule
\cellcolor{black!15}\textbf{Qwen2.5-7B-Instruct}       &\cellcolor{black!15}0.463       &\cellcolor{black!15}43.3\%       &\cellcolor{black!15}0.233       &\cellcolor{black!15}35.0\%       &\cellcolor{black!15}40.0\%       \\
+ Question-Aware Prompt           &0.470      &46.7\%       &0.312       &37.5\%       &43.0\%       \\
+ Knowledge Enhancement  &0.418\phantom{$^{\ast}$} \textcolor{green!60}{$\downarrow$}      &38.3\%\phantom{$^{\ast}$}  \textcolor{green!60}{$\downarrow$}      &0.318       &42.5\%       &40.0\%       \\
+ Self-Improving         &0.557        &55.0\%       &0.291       &37.5\%       &48.0\%       \\
+ Adversarial Finetuning$^{\diamond}$ &0.464       &43.3\%       &0.252       &27.5\%\phantom{$^{\ast}$} \textcolor{green!60}{$\downarrow$}      &37.0\%\phantom{$^{\ast}$} \textcolor{green!60}{$\downarrow$}      \\
+ DoLa Decoding (\textit{low})         &0.420\phantom{$^{\ast}$} \textcolor{green!60}{$\downarrow$}       &43.3\%       &0.283       &35.0\%       & 40.0\%      \\
+ DoLa Decoding (\textit{high})         &0.393\phantom{$^{\ast}$} \textcolor{green!60}{$\downarrow$}       &40.0\%\phantom{$^{\ast}$} \textcolor{green!60}{$\downarrow$}       &0.283       &35.0\%       &38.0\%\phantom{$^{\ast}$} \textcolor{green!60}{$\downarrow$}       \\
+ Anti-DP              &0.344$^{\ast}$ \textcolor{green!60}{$\downarrow$}  &34.5\%$^{\ast}$ \textcolor{green!60}{$\downarrow$}    &0.315       &50.0\%       & 40.8\%      \\
\bottomrule
\end{tabular}
\label{tab:mtg}
\end{table}

\subsection{Mitigation Setup}
We evaluate several mitigation strategies on two currently widely used instruction-tuned LLMs: Llama3.1-8B-Instruct and Qwen2.5-7B-Instruct. For all experiments. For prompt-free strategies, the conversation format remains identical to that used in~\Cref{sec:4.1}. Prompt-based strategies may involve additional text, conversation turns, or both.

We compare conventional strategies (detailed in~\Cref{sec:d}) widely adopted in research on trustworthy AI works, such as those aimed at LLM unbiasing and hallucination elimination. Furthermore, combining insights from RQ1 (~\Cref{sec:4.2}) and RQ2 (~\Cref{sec:5.2}). The considered strategies are:
\begin{itemize}[itemsep=1pt, parsep=0pt]
    \item \textbf{Question-Aware Prompt:} 
    Ask the LLM to be aware of the question, such as ``Please think carefully and cautiously about the question before providing your answer.''
    \item \textbf{Knowledge Enhancement:}
    Provide the LLM with a piece of helpful background knowledge without a direct answer.
    \item \textbf{Self-Improving:} 
    Give the LLM an additional turn to refine its answer with given prompt.
    \item \textbf{Adversarial Finetuning:}
    Finetune the LLM with its unbiased conversations using the LoRA~\citep{lora}.
    \item \textbf{DoLa Decoding:}~\citep{dola}
    Modify the LLM decoding strategy by contrasting the prediction between an early (high or low) layer and the final layer.
    \item \textbf{Anti-DP (Anti-Dual-Process):} 
    Implement a two-phase reasoning intervention. 
    Upon receiving the question, the LLM is first instructed to establish its standard to guide the rethinking. 
    Then, the LLM produces the final answer based on prior thoughts. 
    This approach clearly opposes the automatic dual-process by encouraging integrated, iterative reasoning.
\end{itemize}
\label{sec:6.1}

\subsection{Mitigation Result}
\Cref{tab:mtg} presents the performance of the evaluated mitigation strategies on Llama3.1-8B-Instruct and Qwen2.5-7B-Instruct under the specified sampling setting. 
Mitigation strategies achieved up to a 10\% alleviation in anchoring effects, confirming the failure of elimination, while some reductions are possible. 
Performances are inconsistent, with DoLa Decoding notably demonstrating a more pronounced decrease on semantic tasks for both models. 
Nevertheless, \textbf{Anti-DP} strategy reduces the anchoring effect (except numerical questions on the Qwen model), reinforcing our finding that incorporating an explicit reasoning process during inference helps to erode shallow anchoring.

\section{Conclusion}
In this work, our research investigates the anchoring effect within the context of LLM trustworthiness and cognitive psychology, demonstrating its existence in LLMs and evaluating potential mitigation strategies.
To facilitate this investigation, we introduced \textbf{\textit{SynAnchors}}, a new dataset designed for large-scale studies of the anchoring effect.
By applying the causal tracing methods on the \textbf{\textit{SynAnchors}} dataset, we have contributed to a primary understanding of how LLMs are affected by anchoring hints, i.e., this anchoring pattern is relatively shallow.
Our findings stress the need for further research into cognitive biases within LLMs' inner representation space and powerful techniques to alleviate their impact.
Our study (as discussed in~\Cref{sec:discuss}) also profoundly points out leveraging LLMs' reasoning capabilities as a promising direction for deanchoring.
Through quantifying these biases, we reveal how such cognitive vulnerabilities can undermine LLM reliability in decision-making tasks, where contextual hints may lead to deviations from the objective truth.
Future work will focus on scaling the dataset to a larger collection across comprehensive dimensions to ensure broader generalizability.
We hope this work will make contributions toward unbiased AI.
\clearpage

\bibliography{main}

@article{
firstAE,
author = {Amos Tversky  and Daniel Kahneman },
title = {Judgment under Uncertainty: Heuristics and Biases},
journal = {Science},
volume = {185},
number = {4157},
pages = {1124-1131},
year = {1974},
doi = {10.1126/science.185.4157.1124},
URL = {https://www.science.org/doi/abs/10.1126/science.185.4157.1124},
eprint = {https://www.science.org/doi/pdf/10.1126/science.185.4157.1124}}

@misc{openai2023chatgpt,
  author       = {OpenAI},
  title        = {ChatGPT},
  year         = {2023},
  url          = {https://openai.com/chatgpt},
  note         = {Version GPT-4, Large language model},
}

@article{gpt4o,
  title={Gpt-4o system card},
  author={Hurst, Aaron and Lerer, Adam and Goucher, Adam P and Perelman, Adam and Ramesh, Aditya and Clark, Aidan and Ostrow, AJ and Welihinda, Akila and Hayes, Alan and Radford, Alec and others},
  journal={arXiv preprint arXiv:2410.21276},
  year={2024}
}

@article{rome,
  title={Locating and editing factual associations in GPT},
  author={Meng, Kevin and Bau, David and Andonian, Alex and Belinkov, Yonatan},
  journal={Advances in Neural Information Processing Systems},
  volume={35},
  pages={17359--17372},
  year={2022}
}

@article{Zong2022AnES,
  title={An Experimental Study on Anchoring Effect of Consumers’ Price Judgment Based on Consumers’ Experiencing Scenes},
  author={Yi Zong and Xiaojie Guo},
  journal={Frontiers in Psychology},
  year={2022},
  volume={13},
  url={https://api.semanticscholar.org/CorpusID:246635839}
}

@article{wang2023primacy,
  title={Primacy effect of chatgpt},
  author={Wang, Yiwei and Cai, Yujun and Chen, Muhao and Liang, Yuxuan and Hooi, Bryan},
  journal={arXiv preprint arXiv:2310.13206},
  year={2023}
}

@article{zhang2023towards,
  title={Towards best practices of activation patching in language models: Metrics and methods},
  author={Zhang, Fred and Nanda, Neel},
  journal={arXiv preprint arXiv:2309.16042},
  year={2023}
}

@article{nguyen2024human,
  title={Human bias in AI models? Anchoring effects and mitigation strategies in large language models},
  author={Nguyen, Jeremy K},
  journal={Journal of Behavioral and Experimental Finance},
  volume={43},
  pages={100971},
  year={2024},
  publisher={Elsevier}
}

@article{tversky1974judgment,
  title={Judgment under Uncertainty: Heuristics and Biases: Biases in judgments reveal some heuristics of thinking under uncertainty.},
  author={Tversky, Amos and Kahneman, Daniel},
  journal={science},
  volume={185},
  number={4157},
  pages={1124--1131},
  year={1974},
  publisher={American association for the advancement of science}
}

@article{wong20007300,
  title={Is 7300 m equal to 7.3 km? Same semantics but different anchoring effects},
  author={Wong, Kin Fai Ellick and Kwong, Jessica Yuk Yee},
  journal={Organizational Behavior and Human Decision Processes},
  volume={82},
  number={2},
  pages={314--333},
  year={2000},
  publisher={Elsevier}
}

@article{wilson1996new,
  title={A new look at anchoring effects: basic anchoring and its antecedents.},
  author={Wilson, Timothy D and Houston, Christopher E and Etling, Kathryn M and Brekke, Nancy},
  journal={Journal of Experimental Psychology: General},
  volume={125},
  number={4},
  pages={387},
  year={1996},
  publisher={American Psychological Association}
}

@article{lou2024anchoring,
  title={Anchoring bias in large language models: An experimental study},
  author={Lou, Jiaxu and Sun, Yifan},
  journal={Journal of Computational Social Science},
  volume={9},
  number={1},
  pages={11},
  year={2026},
  publisher={Springer}
}

@article{liu2022trustworthy,
  title={Trustworthy ai: A computational perspective},
  author={Liu, Haochen and Wang, Yiqi and Fan, Wenqi and Liu, Xiaorui and Li, Yaxin and Jain, Shaili and Liu, Yunhao and Jain, Anil and Tang, Jiliang},
  journal={ACM Transactions on Intelligent Systems and Technology},
  volume={14},
  number={1},
  pages={1--59},
  year={2022},
  publisher={ACM New York, NY}
}

@article{brown2020language,
  title={Language models are few-shot learners},
  author={Brown, Tom and Mann, Benjamin and Ryder, Nick and Subbiah, Melanie and Kaplan, Jared D and Dhariwal, Prafulla and Neelakantan, Arvind and Shyam, Pranav and Sastry, Girish and Askell, Amanda and others},
  journal={Advances in neural information processing systems},
  volume={33},
  pages={1877--1901},
  year={2020}
}

@article{gallegos2024bias,
  title={Bias and fairness in large language models: A survey},
  author={Gallegos, Isabel O and Rossi, Ryan A and Barrow, Joe and Tanjim, Md Mehrab and Kim, Sungchul and Dernoncourt, Franck and Yu, Tong and Zhang, Ruiyi and Ahmed, Nesreen K},
  journal={Computational Linguistics},
  volume={50},
  number={3},
  pages={1097--1179},
  year={2024},
  publisher={MIT Press 255 Main Street, 9th Floor, Cambridge, Massachusetts 02142, USA~…}
}

@article{li2023survey,
  title={A survey on fairness in large language models},
  author={Li, Yingji and Du, Mengnan and Song, Rui and Wang, Xin and Wang, Ying},
  journal={arXiv preprint arXiv:2308.10149},
  year={2023}
}

@article{koo2023benchmarking,
  title={Benchmarking cognitive biases in large language models as evaluators},
  author={Koo, Ryan and Lee, Minhwa and Raheja, Vipul and Park, Jong Inn and Kim, Zae Myung and Kang, Dongyeop},
  journal={arXiv preprint arXiv:2309.17012},
  year={2023}
}

@article{shaikh2024cbeval,
  title={CBEval: A framework for evaluating and interpreting cognitive biases in LLMs},
  author={Shaikh, Ammar and Dandekar, Raj Abhijit and Panat, Sreedath and Dandekar, Rajat},
  journal={arXiv preprint arXiv:2412.03605},
  year={2024}
}

@article{meng2022locating,
  title={Locating and editing factual associations in gpt},
  author={Meng, Kevin and Bau, David and Andonian, Alex and Belinkov, Yonatan},
  journal={Advances in neural information processing systems},
  volume={35},
  pages={17359--17372},
  year={2022}
}

@article{wu2024cognitive,
  title={Cognitive LLMs: Towards Integrating Cognitive Architectures and Large Language Models for Manufacturing Decision-making},
  author={Wu, Siyu and Oltramari, Alessandro and Francis, Jonathan and Giles, C Lee and Ritter, Frank E},
  journal={arXiv preprint arXiv:2408.09176},
  year={2024}
}

@article{chapman1999anchoring,
  title={Anchoring, activation, and the construction of values},
  author={Chapman, Gretchen B and Johnson, Eric J},
  journal={Organizational behavior and human decision processes},
  volume={79},
  number={2},
  pages={115--153},
  year={1999},
  publisher={Elsevier}
}

@article{mussweiler2000use,
  title={The use of category and exemplar knowledge in the solution of anchoring tasks.},
  author={Mussweiler, Thomas and Strack, Fritz},
  journal={Journal of personality and social psychology},
  volume={78},
  number={6},
  pages={1038},
  year={2000},
  publisher={American Psychological Association}
}

@article{furnham2011literature,
  title={A literature review of the anchoring effect},
  author={Furnham, Adrian and Boo, Hua Chu},
  journal={The journal of socio-economics},
  volume={40},
  number={1},
  pages={35--42},
  year={2011},
  publisher={Elsevier}
}

@article{jacowitz1995measures,
  title={Measures of anchoring in estimation tasks},
  author={Jacowitz, Karen E and Kahneman, Daniel},
  journal={Personality and social psychology bulletin},
  volume={21},
  number={11},
  pages={1161--1166},
  year={1995},
  publisher={Sage Publications Sage CA: Thousand Oaks, CA}
}

@article{mussweiler2001semantics,
  title={The semantics of anchoring},
  author={Mussweiler, Thomas and Strack, Fritz},
  journal={Organizational behavior and human decision processes},
  volume={86},
  number={2},
  pages={234--255},
  year={2001},
  publisher={Elsevier}
}

@article{mussweiler1999hypothesis,
  title={Hypothesis-consistent testing and semantic priming in the anchoring paradigm: A selective accessibility model},
  author={Mussweiler, Thomas and Strack, Fritz},
  journal={Journal of Experimental Social Psychology},
  volume={35},
  number={2},
  pages={136--164},
  year={1999},
  publisher={Elsevier}
}

@article{strack1997explaining,
  title={Explaining the enigmatic anchoring effect: Mechanisms of selective accessibility.},
  author={Strack, Fritz and Mussweiler, Thomas},
  journal={Journal of personality and social psychology},
  volume={73},
  number={3},
  pages={437},
  year={1997},
  publisher={American Psychological Association}
}

@article{rueda2005training,
  title={Training, maturation, and genetic influences on the development of executive attention},
  author={Rueda, M Rosario and Rothbart, Mary K and McCandliss, Bruce D and Saccomanno, Lisa and Posner, Michael I},
  journal={Proceedings of the National Academy of Sciences},
  volume={102},
  number={41},
  pages={14931--14936},
  year={2005},
  publisher={National Academy of Sciences}
}

@article{munakata2011unified,
  title={A unified framework for inhibitory control},
  author={Munakata, Yuko and Herd, Seth A and Chatham, Christopher H and Depue, Brendan E and Banich, Marie T and O’Reilly, Randall C},
  journal={Trends in cognitive sciences},
  volume={15},
  number={10},
  pages={453--459},
  year={2011},
  publisher={Elsevier}
}

@article{epley2006anchoring,
  title={The anchoring-and-adjustment heuristic: Why the adjustments are insufficient},
  author={Epley, Nicholas and Gilovich, Thomas},
  journal={Psychological science},
  volume={17},
  number={4},
  pages={311--318},
  year={2006},
  publisher={SAGE Publications Sage CA: Los Angeles, CA}
}

@ARTICLE{ieeeaccessLLMAE,
  author={E. O’Leary, Daniel},
  journal={IEEE Intelligent Systems}, 
  title={An Anchoring Effect in Large Language Models}, 
  year={2025},
  volume={40},
  number={2},
  pages={23-26},
  doi={10.1109/MIS.2025.3544939}}

@article{dong2025humanizing,
  title={Humanizing LLMs: A Survey of Psychological Measurements with Tools, Datasets, and Human-Agent Applications},
  author={Dong, Wenhan and Zhao, Yuemeng and Sun, Zhen and Liu, Yule and Peng, Zifan and Zheng, Jingyi and Zhang, Zongmin and Zhang, Ziyi and Wu, Jun and Wang, Ruiming and others},
  journal={arXiv preprint arXiv:2505.00049},
  year={2025}
}

@article{zenren2025rising,
  title={The Rising Threat to Emerging AI-Powered Search Engines},
  author={Luo, Zeren and Peng, Zifan and Liu, Yule and Sun, Zhen and Li, Mingchen and Zheng, Jingyi and He, Xinlei},
  journal={arXiv preprint arXiv:2502.04951},
  year={2025}
}

@book{kahneman2011thinking,
  title={Thinking, fast and slow},
  author={Kahneman, Daniel},
  year={2011},
  publisher={macmillan}
}

@article{He2025AISecuritySurvey,
  author    = {Xinlei He and Guowen Xu and Xingshuo Han and Qian Wang and Lingchen Zhao and Chao Shen and Chenhao Lin and Zhengyu Zhao and Qian Li and Le Yang and Shouling Ji and Shaofeng Li and Haojin Zhu and Zhibo Wang and Rui Zheng and Tianqing Zhu and Qi Li and Chaoxiang He and Qifan Wang and Hongsheng Hu and Shuo Wang and Shi-Feng Sun and Hongwei Yao and Zhan Qin and Kai Chen and Yue Zhao and Hongwei Li and Xinyi Huang and Dengguo Feng},
  title     = {Artificial intelligence security and privacy: a survey},
  journal   = {Science China Information Sciences},
  year      = {2025}
}

@article{fooled2022jcss,
  author       = {Taha Yasseri and
                  Jannie Reher},
  title        = {Fooled by facts: quantifying anchoring bias through a large-scale
                  experiment},
  journal      = {J. Comput. Soc. Sci.},
  volume       = {5},
  number       = {1},
  pages        = {1001--1021},
  year         = {2022},
  url          = {https://doi.org/10.1007/s42001-021-00158-0},
  doi          = {10.1007/S42001-021-00158-0},
  bibsource    = {dblp computer science bibliography, https://dblp.org}
}

@inproceedings{retrHuman,
  author       = {Quan Zhang and
                  Binqi Zeng and
                  Chijin Zhou and
                  Gwihwan Go and
                  Heyuan Shi and
                  Yu Jiang},
  editor       = {Marcelo d'Amorim},
  title        = {Human-Imperceptible Retrieval Poisoning Attacks in LLM-Powered Applications},
  booktitle    = {Companion Proceedings of the 32nd {ACM} International Conference on
                  the Foundations of Software Engineering, {FSE} 2024, Porto de Galinhas,
                  Brazil, July 15-19, 2024},
  pages        = {502--506},
  publisher    = {{ACM}},
  year         = {2024},
  url          = {https://doi.org/10.1145/3663529.3663786},
  doi          = {10.1145/3663529.3663786},
  bibsource    = {dblp computer science bibliography, https://dblp.org}
}

@article{grattafiori2024llama,
  title={The llama 3 herd of models},
  author={Grattafiori, Aaron and Dubey, Abhimanyu and Jauhri, Abhinav and Pandey, Abhinav and Kadian, Abhishek and Al-Dahle, Ahmad and Letman, Aiesha and Mathur, Akhil and Schelten, Alan and Vaughan, Alex and others},
  journal={arXiv preprint arXiv:2407.21783},
  year={2024}
}

@article{yang2024qwen2,
  title={Qwen2. 5 technical report},
  author={Yang, An and Yang, Baosong and Zhang, Beichen and Hui, Binyuan and Zheng, Bo and Yu, Bowen and Li, Chengyuan and Liu, Dayiheng and Huang, Fei and Wei, Haoran and others},
  journal={arXiv preprint arXiv:2412.15115},
  year={2024}
}

@article{deepseek,
  title={Deepseek-r1: Incentivizing reasoning capability in llms via reinforcement learning},
  author={Guo, Daya and Yang, Dejian and Zhang, Haowei and Song, Junxiao and Zhang, Ruoyu and Xu, Runxin and Zhu, Qihao and Ma, Shirong and Wang, Peiyi and Bi, Xiao and others},
  journal={arXiv preprint arXiv:2501.12948},
  year={2025}
}

@article{phi35,
  title={Phi-3 technical report: A highly capable language model locally on your phone},
  author={Abdin, Marah and Aneja, Jyoti and Awadalla, Hany and Awadallah, Ahmed and Awan, Ammar Ahmad and Bach, Nguyen and Bahree, Amit and Bakhtiari, Arash and Bao, Jianmin and Behl, Harkirat and others},
  journal={arXiv preprint arXiv:2404.14219},
  year={2024}
}

@misc{Falcon3,
    title = {The Falcon 3 Family of Open Models},
    url = {https://huggingface.co/blog/falcon3},
    author = {Falcon-LLM Team},
    month = {December},
    year = {2024}
}

@article{yang2025qwen3,
  title={Qwen3 Technical Report},
  author={Yang, An and Li, Anfeng and Yang, Baosong and Zhang, Beichen and Hui, Binyuan and Zheng, Bo and Yu, Bowen and Gao, Chang and Huang, Chengen and Lv, Chenxu and others},
  journal={arXiv preprint arXiv:2505.09388},
  year={2025}
}

@article{jiang2023mistral7b,
  title={Mistral 7B},
  author={Jiang, Albert Q. and Sablayrolles, Alexandre and Mensch, Arthur and Bamford, Chris and Chaplot, Devendra Singh and de las Casas, Diego and Bressand, Florian and Lengyel, Gianna and Lample, Guillaume and Saulnier, Lucile and Lavaud, Lélio Renard and Lachaux, Marie-Anne and Stock, Pierre and Le Scao, Teven and Lavril, Thibaut and Wang, Thomas and Lacroix, Timothée and El Sayed, William},
  journal={arXiv preprint arXiv:2310.06825},
  year={2023},
  url={https://doi.org/10.48550/arXiv.2310.06825}
}

@inproceedings{FromDe,
  author       = {Guy Kaplan and
                  Matanel Oren and
                  Yuval Reif and
                  Roy Schwartz},
  title        = {From Tokens to Words: On the Inner Lexicon of LLMs},
  booktitle    = {The Thirteenth International Conference on Learning Representations,
                  {ICLR} 2025, Singapore, April 24-28, 2025},
  publisher    = {OpenReview.net},
  year         = {2025},
  url          = {https://openreview.net/forum?id=328vch6tRs},
  bibsource    = {dblp computer science bibliography, https://dblp.org}
}

@article{weightDe,
  title={Weight-based Analysis of Detokenization in Language Models: Understanding the First Stage of Inference Without Inference},
  author={Kamoda, Go and Heinzerling, Benjamin and Inaba, Tatsuro and Kudo, Keito and Sakaguchi, Keisuke and Inui, Kentaro},
  journal={arXiv preprint arXiv:2501.15754},
  year={2025}
}

@inproceedings{lora,
  title={LoRA: Low-Rank Adaptation of Large Language Models},
  author={Hu, Edward J and Wallis, Phillip and Allen-Zhu, Zeyuan and Li, Yuanzhi and Wang, Shean and Wang, Lu and Chen, Weizhu and others},
  booktitle={International Conference on Learning Representations}
}

@inproceedings{dola,
  title={DoLa: Decoding by Contrasting Layers Improves Factuality in Large Language Models},
  author={Chuang, Yung-Sung and Xie, Yujia and Luo, Hongyin and Kim, Yoon and Glass, James R and He, Pengcheng},
  booktitle={The Twelfth International Conference on Learning Representations}
}

@inproceedings{mmlu,
  title={Measuring Massive Multitask Language Understanding},
  author={Hendrycks, Dan and Burns, Collin and Basart, Steven and Zou, Andy and Mazeika, Mantas and Song, Dawn and Steinhardt, Jacob},
  booktitle={International Conference on Learning Representations}
}

@inproceedings{truthfulqa,
    title = "{T}ruthful{QA}: Measuring How Models Mimic Human Falsehoods",
    author = "Lin, Stephanie  and
      Hilton, Jacob  and
      Evans, Owain",
    editor = "Muresan, Smaranda  and
      Nakov, Preslav  and
      Villavicencio, Aline",
    booktitle = "Proceedings of the 60th Annual Meeting of the Association for Computational Linguistics (Volume 1: Long Papers)",
    month = may,
    year = "2022",
    address = "Dublin, Ireland",
    publisher = "Association for Computational Linguistics",
    url = "https://aclanthology.org/2022.acl-long.229/",
    doi = "10.18653/v1/2022.acl-long.229",
    pages = "3214--3252"
}

@article{jones2022capturing,
  title={Capturing failures of large language models via human cognitive biases},
  author={Jones, Erik and Steinhardt, Jacob},
  journal={Advances in Neural Information Processing Systems},
  volume={35},
  pages={11785--11799},
  year={2022}
}

@inproceedings{laskar2024systematic,
  title={A systematic survey and critical review on evaluating large language models: Challenges, limitations, and recommendations},
  author={Laskar, Md Tahmid Rahman and Alqahtani, Sawsan and Bari, M Saiful and Rahman, Mizanur and Khan, Mohammad Abdullah Matin and Khan, Haidar and Jahan, Israt and Bhuiyan, Amran and Tan, Chee Wei and Parvez, Md Rizwan and others},
  booktitle={Proceedings of the 2024 Conference on Empirical Methods in Natural Language Processing},
  pages={13785--13816},
  year={2024}
}

@article{wang2025beyond,
  title={Beyond Benchmark: LLMs Evaluation with an Anthropomorphic and Value-oriented Roadmap},
  author={Wang, Jun and Gu, Ninglun and Zhang, Kailai and Zhang, Zijiao and Bao, Yelun and Yang, Jin and Yin, Xu and Liu, Liwei and Liu, Yihuan and Li, Pengyong and others},
  journal={arXiv preprint arXiv:2508.18646},
  year={2025}
}

@inproceedings{priceanchoring2025,
    title = "How Does Cognitive Bias Affect Large Language Models? A Case Study on the Anchoring Effect in Price Negotiation Simulations",
    author = "Takenami, Yoshiki  and
      Huang, Yin Jou  and
      Murawaki, Yugo  and
      Chu, Chenhui",
    editor = "Christodoulopoulos, Christos  and
      Chakraborty, Tanmoy  and
      Rose, Carolyn  and
      Peng, Violet",
    booktitle = "Findings of the Association for Computational Linguistics: EMNLP 2025",
    month = nov,
    year = "2025",
    address = "Suzhou, China",
    publisher = "Association for Computational Linguistics",
    url = "https://aclanthology.org/2025.findings-emnlp.240/",
    doi = "10.18653/v1/2025.findings-emnlp.240",
    pages = "4481--4498",
    ISBN = "979-8-89176-335-7"
}
\bibliographystyle{iclr2026_conference}

\clearpage 

\appendix
\section{Supplementary Background}
\label{sec:supback}
The SAM, proposed by~\citet{strack1997explaining}, challenges the traditional insufficient adjustment hypothesis, which assumes individuals begin at a given anchor and mentally adjust from it. 
Instead, SAM posits that anchoring effects in semantic priming arise from the following cognitive process: first, anchor-induced semantic activation facilitates access to semantically consistent information; second, this information becomes selectively accessible during judgment. 
Crucially, SAM emphasizes that anchoring can occur even when the anchor value falls within a plausible range and is not dependent on its extremity~\citep{chapman1999anchoring,mussweiler1999hypothesis}.

This model is motivated by empirical findings that contradict the adjustment-based explanation. 
Through incentives or warnings, anchoring is not reliably reduced~\citep{epley2006anchoring}, and anchoring effects remain stable regardless of the anchor's salience or contextual features~\citep{mussweiler2000use}. 
These observations support the idea that anchoring is primarily driven by automatic, hypothesis-consistent semantic activation, rather than deliberate correction.
SAM thus reframes anchoring not as a failure of cognitive effort, but as an automatic cognitive process shaped by how humans retrieve and evaluate information with limited cognitive resources. 

\section{\textbf{\textit{SynAnchors}}}
\label{sec:b}
\subsection{Dataset Samples and Statistics}
Dataset quality is crucial for conducting meaningful research. Below are samples from our dataset, \textbf{\textit{SynAnchors}}, showcasing the variety and high quality of the semantic and numerical tasks, with two examples provided for each. As for the anchor values, the high/low anchor values of semantic anchoring questions are distinctive, ranging from 0.5 to 2.0 times the true value.
\begin{tcolorbox}[
  enhanced,
  breakable,
  colback=black!3!white,
  colframe=black!60!white,
  fonttitle=\bfseries,
  title={Semantic Question Sample 1},
  coltitle=white,
  colbacktitle=black!60!white,
  boxrule=0.8pt,
  top=4pt,
  bottom=4pt,
  left=5pt,
  right=5pt,
]
\textbf{Anchoring Item}: \textit{Annual water consumption of global avocado production} \\ \textbf{Question}: \textit{Considering the intensive irrigation needs of avocado trees across major producing countries, what would you estimate is the total volume of water used each year globally for commercial avocado cultivation? (in billion cubic meters)} \\ \textbf{Anchor Text:} Is it higher or lower than \{\} billion cubic meters? \\  \textbf{True Value}: 6.9 \\ \textbf{Low Anchor}: 3.65\\ \textbf{High Anchor}: 10.85 \\ \textbf{Topic}: Resource Consumption
\end{tcolorbox}

This sample presents a factual question on a real-world topic, designed to be unfamiliar or scarcely represented in typical LLM training corpora. This characteristic is crucial for evaluating the anchoring effect, as it ensures that the LLM's response relies less on pre-existing knowledge and more on the provided anchor.

\begin{tcolorbox}[
  enhanced,
  breakable,
  colback=black!3!white,
  colframe=black!60!white,
  fonttitle=\bfseries,
  title={Semantic Question Sample 2},
  coltitle=white,
  colbacktitle=black!60!white,
  boxrule=0.8pt,
  top=4pt,
  bottom=4pt,
  left=5pt,
  right=5pt,
]
\textbf{Anchoring Item}: \textit{Willingness to pay for a movie ticket} \\ \textbf{Question}: \textit{If you were going to see a standard 2D showing of a new release movie at your local cinema, how much would you realistically be willing to pay for one adult ticket (in USD)?} \\ \textbf{Anchor Text:} Is it higher or lower than \{\} USD? \\  \textbf{True Value}: 15.5 \\ \textbf{Low Anchor}: 12.13\\ \textbf{High Anchor}: 21.22 \\ \textbf{Topic}: Willingness to Pay
\end{tcolorbox}

This sample, in contrast to the previous factual query, delves into a daily-life judgment scenario. It is designed to test the LLM's anchoring effect from a human-like perspective, simulating decision-making processes found in everyday human contexts.

\begin{tcolorbox}[
  enhanced,
  breakable,
  colback=black!3!white,
  colframe=black!60!white,
  fonttitle=\bfseries,
  title={Numerical Question Sample 1},
  coltitle=white,
  colbacktitle=black!60!white,
  boxrule=0.8pt,
  top=4pt,
  bottom=4pt,
  left=5pt,
  right=5pt,
]
\textbf{Anchoring Item}: Number of trees in the Amazon rainforest \\ \textbf{Question}: What is the estimated number of trees in the Amazon rainforest? (in billions) \\ \textbf{Anchor Text:} The number of socks in your drawer: \{\}\\  \textbf{True Value}: 390 \\ \textbf{Anchor Value}: 939\\ \textbf{Topic}: Quantity
\end{tcolorbox}

\begin{figure*}[!t]
    \centering
    \begin{minipage}{0.48\linewidth}
        \centering
    \includegraphics[width=1\linewidth]{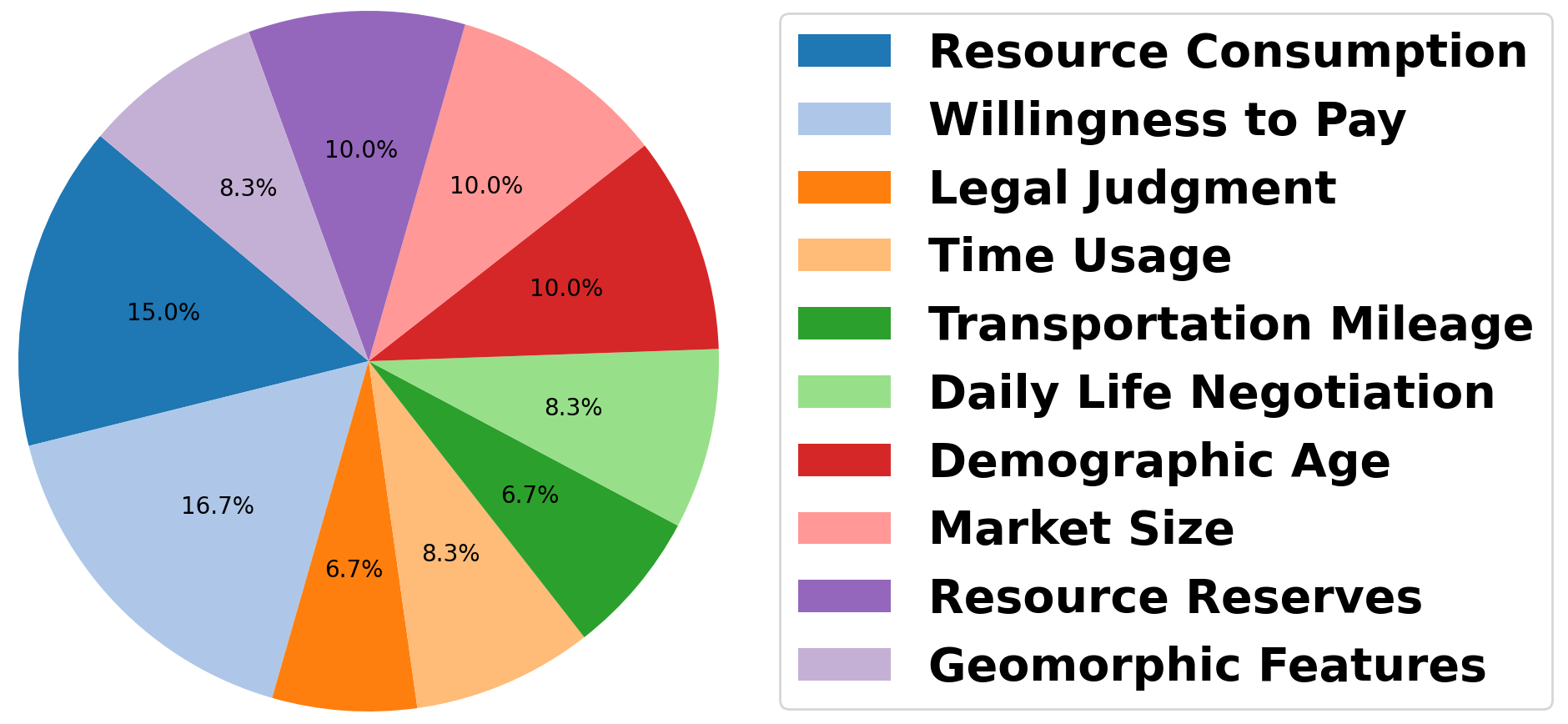}
    \caption{Categories of topic in semantic questions.}
    \label{fig:semantic_topics}
    \end{minipage}
    \hfill
    \begin{minipage}{0.48\linewidth}
        \centering
    \includegraphics[width=0.88\linewidth]{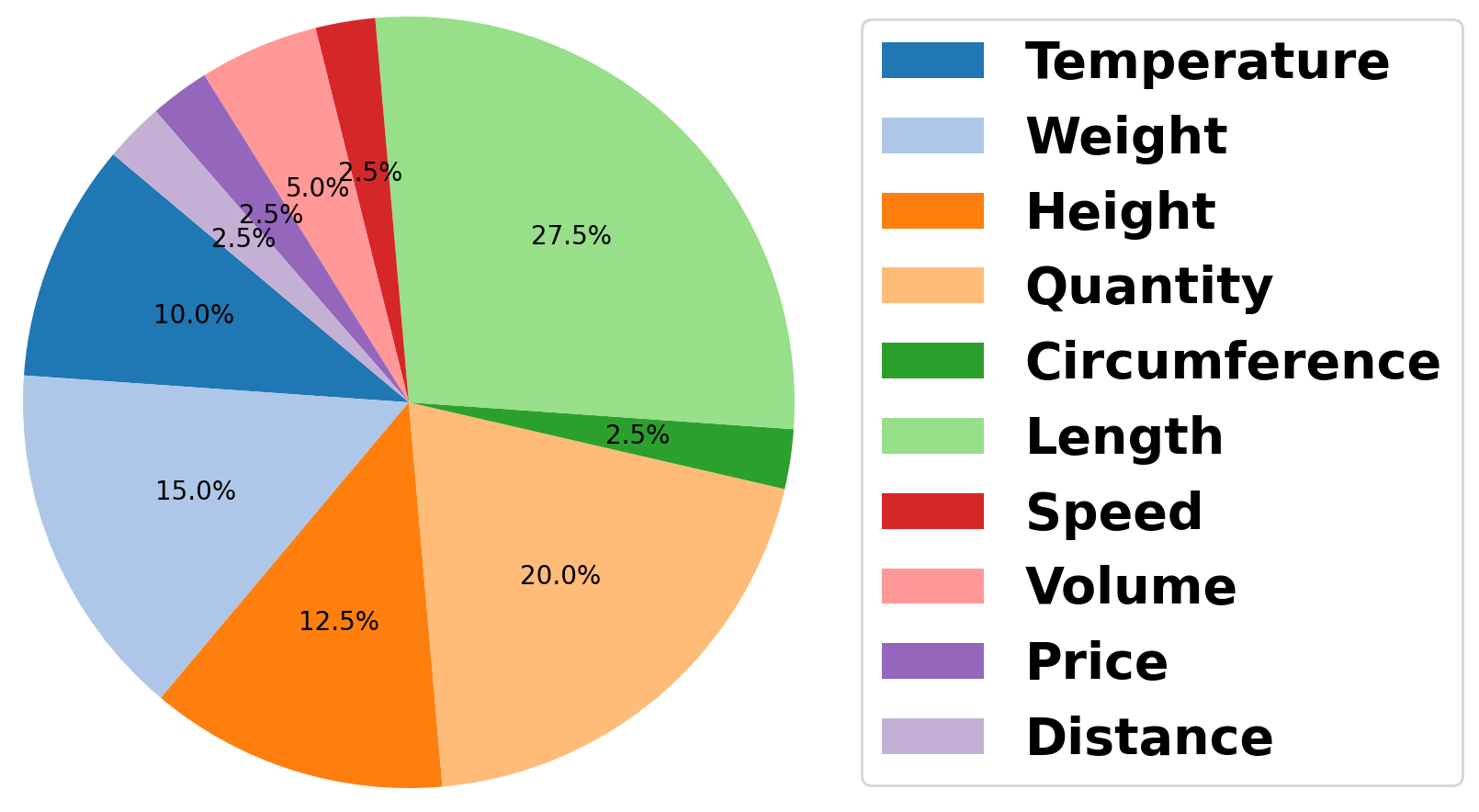}
    \caption{Categories of topic in numerical questions.}
    \label{fig:numerical_topics}
    \end{minipage}
\end{figure*}

\Cref{fig:semantic_topics} illustrates the topic diversity of semantic anchoring questions across various subfields of factual knowledge and daily life decision-making, making it representative for testing anchoring effects in a human-like style.

\Cref{fig:numerical_topics} showcases the topic diversity of numerical anchoring questions concerning quantitative measurements, covering both discrete and continuous factual data. This makes the dataset representative of numerical judgment scenarios typically encountered by humans.
This sample from the \textbf{\textit{SynAnchors}} dataset represents a question on a discrete quantitative measurement. It probes real-world factual knowledge about the estimated number of trees in the Amazon rainforest, making it a truth-verifiable query with a true value of 390 billion.

\begin{tcolorbox}[
  enhanced,
  breakable,
  colback=black!3!white,
  colframe=black!60!white,
  fonttitle=\bfseries,
  title={Numerical Question Sample 2},
  coltitle=white,
  colbacktitle=black!60!white,
  boxrule=0.8pt,
  top=4pt,
  bottom=4pt,
  left=5pt,
  right=5pt,
]
\textbf{Anchoring Item}: The Eiffel Tower's shrinkage in winter \\ \textbf{Question}: How much does the Eiffel Tower shrink in winter due to cold (cm)? \\ \textbf{Anchor Text:} The number of notifications on your phone: \{\}\\  \textbf{True Value}: 15 \\ \textbf{Anchor Value}: 514\\ \textbf{Topic}: Height
\end{tcolorbox}

This sample features a question centered on continuous quantitative measurement and real-world knowledge. The query is truth-verifiable and designed to be unfamiliar yet reasonable for LLMs. 

Note that for each numerical question, the anchor text and its corresponding anchor value are varied to investigate the influence of arbitrary numerical primes on estimation.

\begin{figure*}[!t]
    \centering
    \begin{minipage}{0.48\linewidth}
        \centering
    \includegraphics[width=1\linewidth]{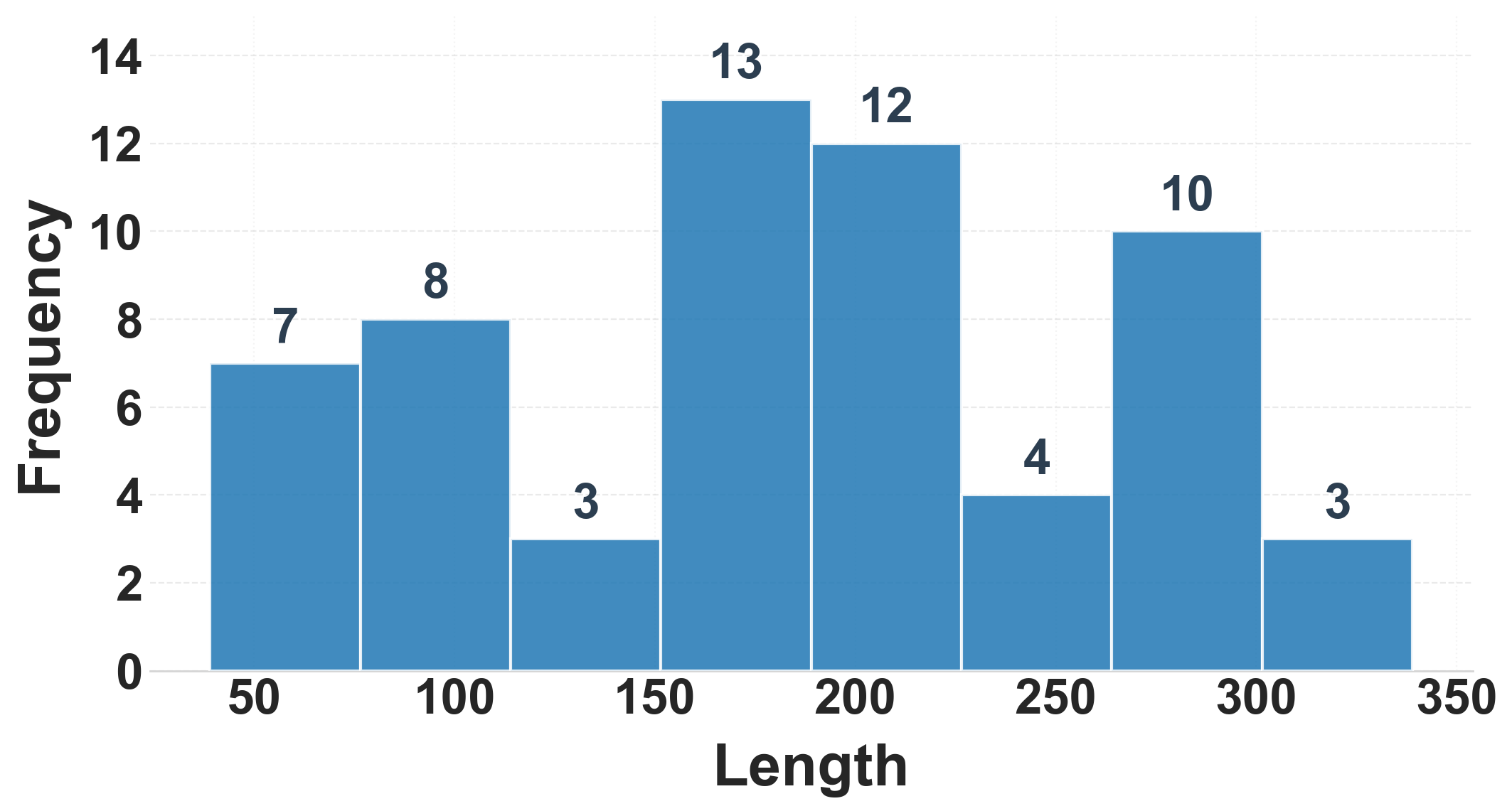}
    \caption{Length distribution of semantic questions.}
    \label{fig:semantic_length}
    \end{minipage}
    \hfill
    \begin{minipage}{0.48\linewidth}
        \centering
    \includegraphics[width=1\linewidth]{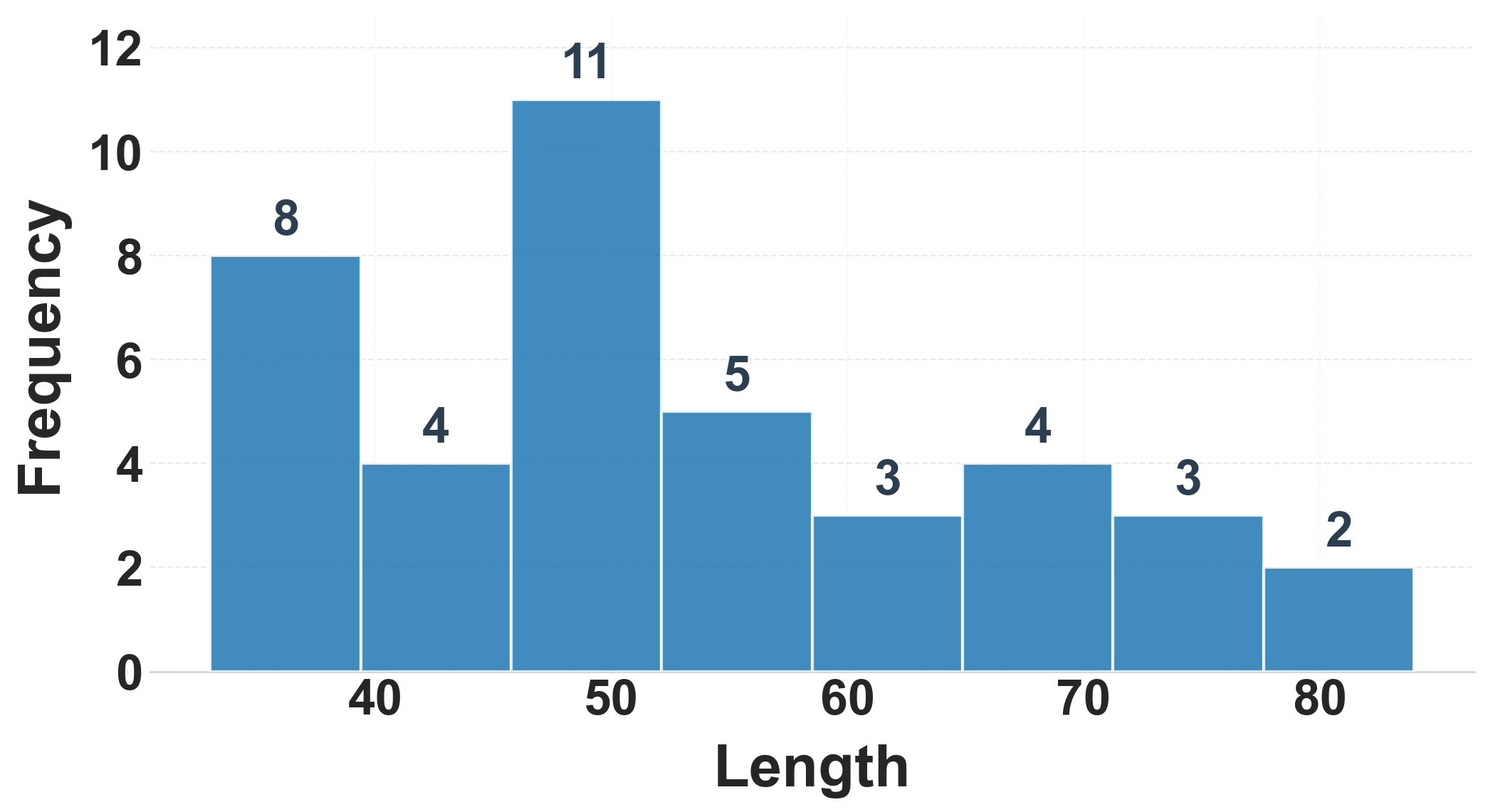}
    \caption{Length distribution of numerical questions.}
    \label{fig:numerical_length}
    \end{minipage}
\end{figure*}

\Cref{fig:semantic_length} and~\Cref{fig:numerical_length} illustrate the diversity of question length in both semantic and numerical anchoring questions. This variation ensures linguistic diversity, thereby enabling the activation of diverse semantic features through varied linguistic contexts.


\subsection{Human-LLM-Loop Data Curation}
\textbf{Initial Generation:} In this process, we begin by developing several foundational seed questions. These are carefully crafted by human annotators to establish a broad spectrum of topics and question structures. For instance, seed questions span both real-world factual knowledge (e.g., ``Number of trees in the Amazon rainforest") and daily-life judgment scenarios (e.g., ``Willingness to pay for a movie ticket"). This initial set serves as the blueprint for the structural and topical patterns we aim to replicate in the larger generated pool. We then leverage the powerful LLM DeepSeek-R1 to generate an extensive pool of candidate questions. This process yields an initial pool of approximately 10 raw candidate questions. A representative prompt used for this generation step is: 
\begin{tcolorbox}[
  enhanced,
  breakable,
  colback=black!3!white,
  colframe=black!60!white,
  fonttitle=\bfseries,
  title={Initial Generation Prompt},
  coltitle=white,
  colbacktitle=black!60!white,
  boxrule=0.8pt,
  top=4pt,
  bottom=4pt,
  left=5pt,
  right=5pt,
]
\textbf{Instruction}: \textit{Please generate 10 questions similar in structure, complexity, and topic to the following examples. Ensure questions are concise, verifiable, and relate to either real-world facts or daily life human judgments.} \\ \textbf{Example 1}: (Resource Reserves) What is the estimated total global silver reserve in 2024? (in metric tons) \\ \textbf{Example 2}: (Willingness to Pay) How much would you realistically be willing to pay for a takeout meal (in USD)? 
\end{tcolorbox}

\textbf{Human Curation and Filtering}: 
The generated pool of candidate questions undergoes meticulous human review and filtering to ensure adherence to our quality principles. 
Annotators meticulously review each candidate question with the following criteria:
\begin{enumerate}
    \item \textbf{Verifiability with a True Value}: Questions requiring factual answers should be verifiable through reliable public sources (e.g., academic papers, reputable databases, government reports). Questions related to subjective judgments (e.g., willingness to pay, legal judgments) should have a reasonably acceptable value aligned with common cases.
    \item \textbf{Unfamiliarity with Common LLMs}: A critical criterion is to select questions unlikely to be present in the explicit training corpora of common LLMs. For instance, while questions about the Eiffel Tower's height are common, its shrinkage in winter is a specific detail less likely to be memorized, forcing the LLM to process anchors more directly. This ensures that observed biases are due to anchoring instead of pre-existing knowledge.
    \item \textbf{Balance of Topics}: Questions are categorized into diverse semantic topics (~\Cref{fig:semantic_topics}) and numerical categories (~\Cref{fig:numerical_topics}), ensuring coverage of real-world knowledge and daily life decision-making. This diversity enhances the dataset's representativeness for human-like judgment scenarios.
    \item \textbf{Diversity of Anchoring Items}: The specific anchoring items, as the core object of the anchoring question (e.g., weight of a cat, height of Everest) are diversified.
    \item \textbf{Linguistic Diversity}: Questions are assessed for linguistic variety in terms of phrasing, linguistic context, and length. This diversity, as presented in ~\Cref{fig:semantic_length} and ~\Cref{fig:numerical_length}, is crucial for activating diverse semantic features and creating varied linguistic situations for the LLMs.
\end{enumerate}
Candidate questions that seriously deviate from the above criteria are discarded.

\textbf{True Value Determination}:
For each filtered question, human annotators meticulously determine its true value through a systematic process of web searching and cross-verification. For factual questions (e.g., ``Number of trees in the Amazon rainforest"), annotators use multiple reputable sources (e.g., scientific databases, government environmental reports) to ascertain the most accurate and widely accepted value. For judgment-based questions (e.g., ``Willingness to pay for a movie ticket"), true values are often derived from established survey data or market research where available, representing a consensus on human judgment. This multi-source verification ensures the robustness and accuracy of our true value.

\textbf{Iterative Refinement}:
The final stage involves an iterative refinement loop between human annotators and the LLM's output. Human annotators provide structured feedback on the generated questions, pointing out specific areas for improvement (e.g., unverifiability, insufficient unfamiliarity, monotonous expressions, and topical imbalance). This feedback is then used to refine subsequent generation cycles, enabling the model to learn and improve its output quality. This cycle continues until the desired volume of high-quality questions is achieved, meeting all the aforementioned criteria for verifiability, unfamiliarity, and diversity. This approach ensures that the dataset is not only extensive but also meticulously aligned with our research objectives.

\section{Causal Tracing Details}
\label{sec:casual_tracing_details}
To be specific, we focus on reverse engineering the LLM to explore certain tokens' (and their relative hidden states') role in certain layers, especially tokens carrying anchoring hints. We conduct a detailed causal tracing analysis to explore how the anchoring effect manifests in large language models (LLMs). 
\begin{enumerate} 
     \item \textbf{Clean Run}: In the clean run, we pass the whole question $x$ into the model and collect all hidden activations $\{h(l)_i | i \in [1, T], l \in [1, L]\}$ at each layer $l$;
     \item \textbf{Corrupted Run}: The corrupted run involves obfuscating the key components of the input, which are regarded as region of interest (ROI) tokens. These tokens are usually the important tokens for LLMs' response generation, and generally are tokens about the subject/object. Considering our research goal,  tokens about anchoring hints are also included. To be specific, we manipulate these tokens by adding noise to their input embeddings $h(0)_i$ with noise $\epsilon$, to simulate a “corrupted” input. In this run, we save the model output logits of the newest next token prediction (which is the numerical answer to the final questions).
     \item \textbf{Restored Run}: In the restored run, we patched the clean activation from the clean run back into the corrupted state. Specifically, at a particular corrupted token's corresponding hidden states $h(l)_{i}^*$, we restore the activation to the value obtained in the clean run. The restored activation allows the model to recover some of its prior performance. The effectiveness of this patching operation is measured by the difference in LLM's output (i.e., the next token prediction) between the corrupted and restored runs.
\end{enumerate}
The recovery effect of certain tokens in certain layers represents their original significance in the model's inner workings, which is expressed by the Kullback-Leibler (KL) divergence difference between clean runs with respect to corruption runs and clean runs with respect to restore runs of the newest generated token's probability distributions.
In our case, the clean run provides a reference, the corrupted run introduces the anchoring manipulation of all tokens we are interested in, and the restoration run helps us gauge how much the model relies on specific tokens (such as those corresponding to the anchor values) to generate the correct answer. By comparing the model's performance across these different runs, we can quantify the total effect (TE) and indirect effect (IE) of the anchoring hint on the model’s answer.
Mathematically, it is ($P_{\text{cl}}$, $P_{*}$, and $P_{\text{pt}}$ are respectively the distributions of clean, corrupted, and restored runs):
\begin{equation}
    \Delta D_{KL} = D_{\text{KL}}(P_{\text{cl}} \parallel P_{*}) -  D_{\text{KL}}(P_{\text{cl}} \parallel P_{\text{pt}}). \notag
\end{equation}
After collecting all the Kullback-Leibler divergence differences of corresponding hidden states, it allows us to visualize the significance of different hidden states.

\section{Causal Tracing Results on Numerical Questions}
\label{ctr_nq}
As for numerical questions, target tokens are:
\begin{itemize}[itemsep=1pt, parsep=0pt]
\item \textbf{\texttt{a\_subj\_1st}}, \textbf{\texttt{a\_subj\_last}}: The first and last tokens of the subject in \texttt{<anchor\_text>}.
\item \textbf{\texttt{a\_num\_1st}}, \textbf{\texttt{a\_num\_last}}: The first and last tokens of the numerical expression in \texttt{<anchor\_text>}.
\item \textbf{\texttt{a\_avg}}: Average significance over all tokens in \texttt{<anchor\_text>}.
\item \textbf{\texttt{q\_subj\_1st}}, \textbf{\texttt{q\_subj\_last}}: The first and last tokens of the subject in \texttt{question}.
\item \textbf{\texttt{else\_avg}}, \textbf{\texttt{else\_avg2}}, \textbf{\texttt{all\_avg}}: Average significance over the tokens not marked by any special role (for different stages or components) of \textbf{System Prompt}, \textbf{Question}, and all tokens, respectively.
\end{itemize}
With the prepared means in the main text and above, we derive the result in~\Cref{fig:n_llama} here.
\begin{figure}[!htbp]
    \centering
    \includegraphics[width=1.0\linewidth]{n_llama.pdf}
    \caption{Causal tracing on attention (red) and FFN (green) modules of LLama-3.1-8B-Instruct about numerical anchoring questions. The X-axis represents the layer index of the model (32 layers). The Y-axis is the ROI tokens.}
    \label{fig:n_llama}
\end{figure}
As shown in~\Cref{fig:n_llama}, tokens of irrelevant numerical anchoring hints contribute to some extent to both attention and FFN modules. These contributions are still lighter than the subject tokens in question and rarely demonstrate salience in higher layers. 

\section{Details of Mitigation Strategies}
Their concrete implementations are demonstrated below.
\label{sec:d}
\paragraph{Question-Aware Prompt.}
The Question-Aware Prompt strategy aims to prime the Large Language Model (LLM) to engage in a more deliberate and cautious processing of the input question. The additional hint added to the prompt is as follows:
\begin{tcolorbox}[
  enhanced,
  breakable,
  colback=black!3!white,
  colframe=black!60!white,
  fonttitle=\bfseries,
  title={Question-Aware Prompt},
  coltitle=white,
  colbacktitle=black!60!white,
  boxrule=0.8pt,
  top=2pt,
bottom=2pt,
left=2pt,
right=2pt,
]
\textbf{Additional Prompt:} Interpret the question carefully and think cautiously.
\end{tcolorbox}

\paragraph{Knowledge Enhancement.}
The Knowledge Enhancement strategy investigates whether providing relevant, non-answer-revealing background information can help LLMs overcome anchoring effects. 
For each question, a concise piece of helpful background knowledge relevant to the question's topic is appended to the prompt. This background information is carefully tailored to avoid revealing the true answer directly. For example:
\begin{tcolorbox}[
  enhanced,
  breakable,
  colback=black!3!white,
  colframe=black!60!white,
  fonttitle=\bfseries,
  title={Knowledge Enhancement Prompt},
  coltitle=white,
  colbacktitle=black!60!white,
  boxrule=0.8pt,
  top=2pt,
bottom=2pt,
left=2pt,
right=2pt,
]
\textbf{Question:} How many billion pieces of plastic packaging waste do UK homes discard annually? \\
\textbf{Additional Prompt:} You will be provided with some background knowledge, which starts with notion [Background knowledge]. [Background knowledge]: There are approximately 27.8 million households in the UK.
\end{tcolorbox}
This background knowledge offers helpful auxiliary information in calculating or estimating the final answer. 

\paragraph{Self-Improving.} This approach is inspired by self-reflection techniques used in other LLM contexts to enhance output quality.
\begin{tcolorbox}[
  enhanced,
  breakable,
  colback=black!3!white,
  colframe=black!60!white,
  fonttitle=\bfseries,
  title={Self-Improving Conversation},
  coltitle=white,
  colbacktitle=black!60!white,
  boxrule=0.8pt,
  top=2pt,
bottom=2pt,
left=2pt,
right=2pt,
]
\textbf{Original Question:}[Original Question]\\
\textbf{Previous Answer}: [Previous Answer]\\
\textbf{Prompt of Additional Turn:} Please rethink the above answer and give a more accurate answer. 
\end{tcolorbox}

\paragraph{Adversarial Finetuning.}
Adversarial Finetuning aims to mitigate anchoring bias by training the LLM on the unbiased questions of the dataset. We employ the LoRA (Low-Rank Adaptation) method for parameter-efficient finetuning of the two selected standard light models:  Llama-3.1-8B-Instruct and Qwen2.5-7B-Instruct. 
The LoRA finetuning configuration is:
(1) LoRA Rank (r): 64. (2) LoRA Alpha: 128. (3) LoRA Dropout: 0.1. (4) Optimizer: AdamW. (5) Learning Rate: $5e^{-5}$. (6) Number of Epochs: 3. (7) Training Batch Size: 8. 
The calculation of adversarial finetuning's performance is slightly different from usual groups: The questions are split into the training set and a testing set. Questions contained in the training set are frozen, reserving their results in the baseline, while the test set is evaluated. The results of the two split sets are then merged into the overall performance.

\paragraph{DoLa.}
DoLa (Decoding by Contrasting Layers) is a decoding-time modification strategy within the LLM, contrasting the activation patterns between an early layer and the final layer. This aims to reduce reliance on potentially shallow or heuristic-driven activations that might be more susceptible to anchoring. 

\paragraph{Anti-DP}
From a cognitive-aware mitigation perspective, we implement \textbf{Anti-DP} as a two-phase reasoning intervention. This strategy is built upon the dual-process (DP) theory of cognition, aiming to steer the LLM towards System 2-like reasoning to overcome System 1-like biases induced by anchors. Anti-DP operationalizes a structured cognitive-inspired strategy within the LLM's generation process. It forces the model through an initial analytical phase to establish internal guidelines before formulating its final response, in a limited reasoning budget of up to 128 tokens.

\begin{tcolorbox}[
  enhanced,
  breakable,
  colback=black!3!white,
  colframe=black!60!white,
  fonttitle=\bfseries,
  title={Anti-DP Conversation},
  coltitle=white,
  colbacktitle=black!60!white,
  boxrule=0.8pt,
  top=2pt,
bottom=2pt,
left=2pt,
right=2pt,
]
\textbf{Standard Conversation:}[Standard Conversation]\\
\textbf{Anti-DP Prompt:} Before accepting any given initial reference value, identify your independent criteria for the answer to this question. Ask: How would I assess this if no reference value is provided? What objective standards exist outside the given information of the question? Establish your own criteria first, then rethink the answer using your independent criteria through unbiased reasoning.\\
\textbf{Anti-DP Reasoning:} [Anti-DP Reasoning]  \\
\textbf{Final Prompt:} Please give a more accurate answer based on your previous thoughts.
\end{tcolorbox}

\section{Discussion}
\label{sec:discuss}
Our results show that anchoring effects are prominent in LLMs (RQ1), that enhancing reasoning may offer a path to reduce such shallow biases (RQ2), and that current mitigation methods fail to fully address them (RQ3). 
The \textbf{Anti-DP} intervention introduces an alternative activation path that may interfere with or override the anchor-primed semantic activation, akin to inhibitory control mechanisms~\citep{rueda2005training,munakata2011unified} observed in human cognition, allowing the model to suppress automatic responses in favor of goal-directed behavior.
This effect parallels human cognition: in dual-process theories~\citep{kahneman2011thinking}, intuitive judgments (System 1) are prone to anchoring, while reflective reasoning (System 2) can correct such biases. 
Similarly, reasoning prompts in LLMs may functionally resemble System 2 by interrupting default predictions and guiding controlled generation.

At a low semantic level, semantic cues like ``higher" or ``lower" exert a stronger influence on the model’s output, shaping a shallow form of selective accessibility centered around the anchor. This mirrors SAM’s central claim in humans: anchoring stems from the selective activation of semantically consistent information, rather than from deliberate adjustment. This activation is automatic and pre-reflective, echoing the effortless nature of human heuristic processing.
Building on this perspective, we tested whether prompt-based interventions could disrupt these semantic pathways. 
Prior psychological research suggests that in comparative judgment, humans often engage in hypothesis testing to the anchor, and that redirecting attention to diagnostic feature of the task instead of comparison of anchor and target can reduce anchoring~\citep{chapman1999anchoring}. The Anti-DP intervention prompts models toward task-relevant reasoning, thereby overriding anchor-primed activation to some extent.

These findings suggest that anchoring in LLMs is not a static or hardcoded trait, but an emergent property of context-sensitive processing—one that can be modulated externally. SAM thus offers a useful theoretical reference point for both interpreting anchoring effects and designing effective interventions, underscoring the value of integrating psychological theory into LLM research.
\end{document}